\documentclass[10pt,twocolumn,twoside,journal]{IEEEtran} 
\usepackage{lipsum}
\usepackage{cite}
\usepackage{amsmath, amssymb,amsfonts}
\usepackage{algorithmic}
\usepackage{graphicx}
\usepackage{textcomp}
\usepackage{hyperref}
\usepackage{enumerate}
\hypersetup{hidelinks,colorlinks=true,allcolors=red,pdfstartview=Fit,breaklinks=true}

\begin{document}
	
	\title{A Survey on Passing-through Control of Multi-Robot Systems in Cluttered Environments}
	
	\author{Yan Gao, Chenggang Bai, Quan Quan
	\thanks{Yan Gao, Chenggang Bai and Quan Quan are with the School of Automation Science and Electrical Engineering, Beihang University, Beijing 100191, P. R. China (email: buaa\_gaoyan@buaa.edu.cn; bcg@buaa.edu.cn; qq\_buaa@buaa.edu.cn)}}
	
	
\maketitle

\begin{abstract}
This survey presents a comprehensive review of various methods and algorithms related to passing-through control of multi-robot systems in cluttered environments. Numerous studies have investigated this area, and we identify several avenues for enhancing existing methods. This survey describes some models of robots and commonly considered control objectives, followed by an in-depth analysis of four types of algorithms that can be employed for passing-through control: leader-follower formation control, multi-robot trajectory planning, control-based methods, and virtual tube planning and control. Furthermore, we conduct a comparative analysis of these techniques and provide some subjective and general evaluations.
\end{abstract}


\section{Introduction}

\subsection{Background}

In recent years, multi-robot systems have been an important topic of robotic research. Multi-robot systems now have a level of sophistication thanks to research developments, and are becoming increasingly attractive for a variety of complex applications, including sensing, mapping, search and rescue, and some military tasks. In order to achieve these applications, it is necessary for the multi-robot system to have the ability to move in a complex and cluttered environment, and reach the appointed destination. In this process, each robot should avoid collisions with obstacles and other robots to keep safety.

\begin{figure}[!t]
	\begin{centering}
		\includegraphics[width=\columnwidth]{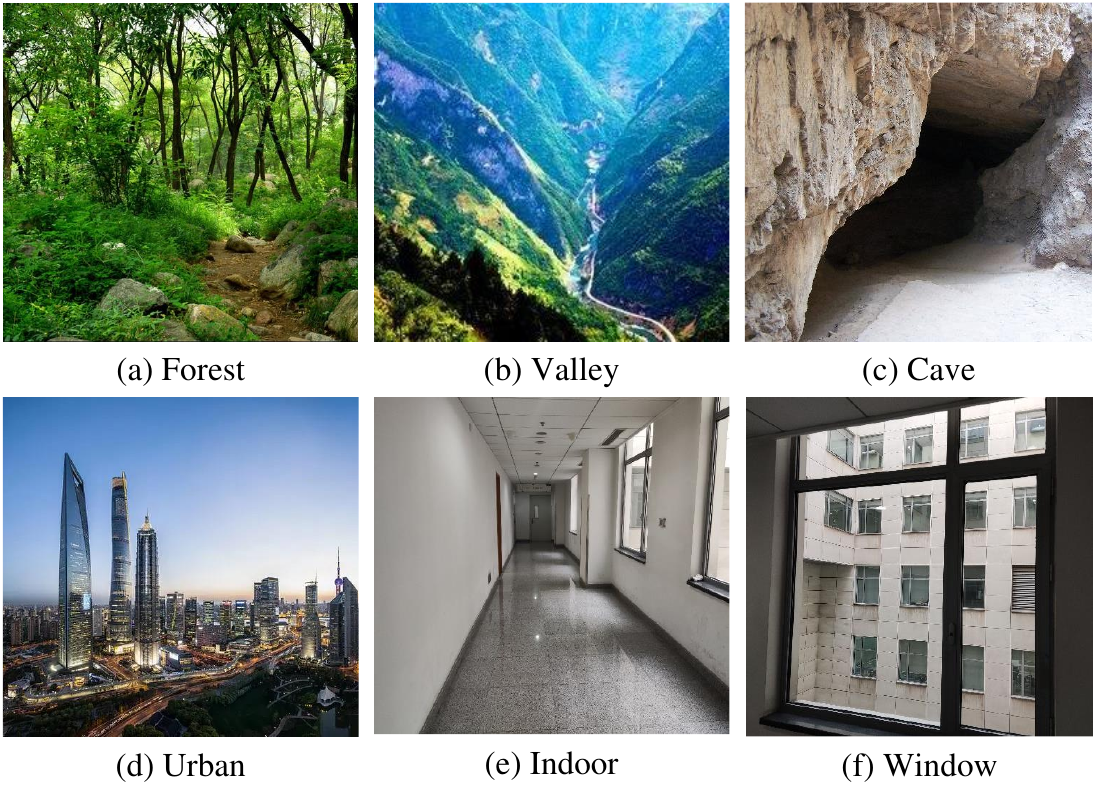}
		\par \end{centering}
	\caption{Examples of cluttered environments.}
	\label{Cluttered}
\end{figure}

In this survey, we refer to traversing the cluttered environment as a ``passing-through'' process. As shown in Figure \ref{Cluttered}, the cluttered environment here includes forests, valleys, indoor space, narrow waterways, etc. Moving within a narrow corridor, through a window or a doorframe is also a very common scenario. Besides, the traffic management of automated road vehicles and air traffic management of unmanned aerial vehicles can also be viewed as a specific ``passing-through'' application. The robot in the multi-robot system can be a mobile ground vehicle, a multicopter, a fixed-wing unmanned aerial vehicle (UAV),  an unmanned surface vehicle (USV), an unmanned underwater vehicle (UUV), etc. Different kinds of robots correspond to different mathematical models, which will be described in detail in the following.

\subsection{Algorithms for Passing-through Control}
\begin{figure*}[!t]
	\begin{centering}
		\includegraphics[width=0.9\textwidth]{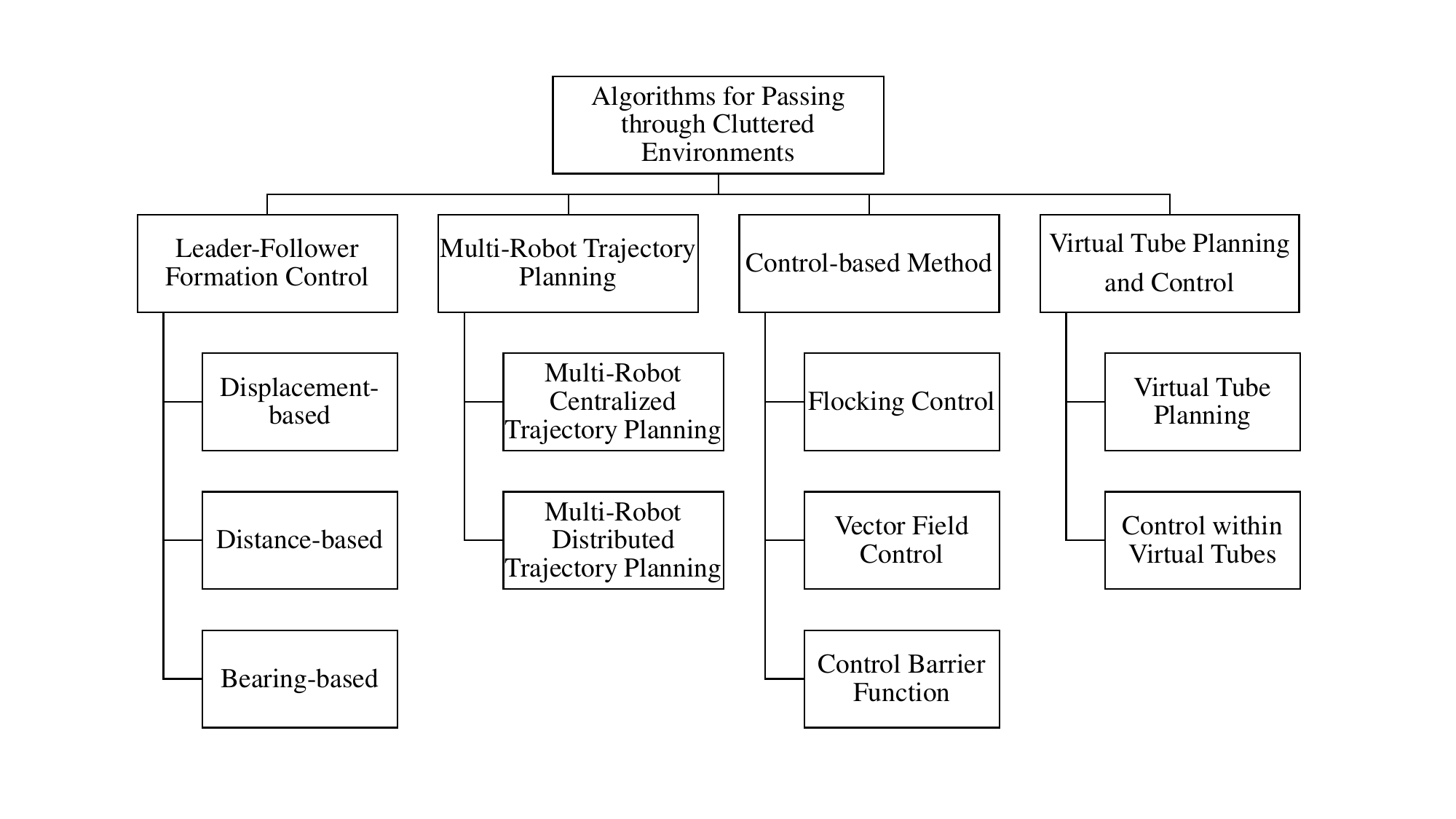}
		\par \end{centering}
	\caption{Diagram of algorithms for passing through cluttered environments.}
	\label{Diagram}
\end{figure*}

Many successful algorithms have been put forward for the passing-through control of a multi-robot system in a cluttered environment. As shown in Figure \ref{Diagram}, the typical algorithms can mainly be classified into four types: leader-follower formation control, multi-robot trajectory planning, control-based methods, virtual tube planning and control. 

The research area of multi-robot systems places considerable emphasis on the topic of formation control. The existing formation control strategies can be roughly classified into two categories: the leader-follower approach and the virtual structure approach. In some literature, the control strategy for the formation also includes the behavior-based approach \cite{oh2015survey,jiang2022bibliometric}, which is considered to belong to control-based methods in this survey. Compared to the virtual structure approach, the leader-follower approach is more suitable for guiding multiple robots to pass through a cluttered environment. The reason is that the virtual structure approach is not flexible and has a poor ability in obstacle avoidance \cite{do2021formation}. The leader-follower approach is the most common method in formation control due to its simple control structure. In this approach, a robot in the multi-robot system is nominated as a leader, and the other robots are considered as followers. In most cases, leader-follower formation control algorithms can be implemented in a distributed manner. However, scalability and adaptability present significant challenges. For instance, when the number of robots is scaled up to hundreds or thousands, the physical size of the formation becomes too large for practical feasibility. Furthermore, when some robots need to change their locations, it may cause chaos in the formation and the leader-follower formation controller will become very complex.

Multi-robot trajectory planning involves generating collision-free trajectories with higher-order continuity for each robot either in a centralized or a distributed fashion, and then guiding each robot to follow its corresponding trajectory. In contrast to velocity or acceleration commands in formation control, higher-order trajectories, such as cubic splines \cite{chen2020dynamic}, Bezier curves \cite{askari2016new}, and B-splines \cite{zhou2020robust}, offer improved control performance. To reach a specified target point, each robot in the multi-robot system should first find a discrete geometric path in the global map before locally optimizing the path to produce a feasible trajectory that avoids obstacles and other robots' trajectories. If the multi-robot trajectory planning is centralized, a designated central node computes trajectories for all robots and then sends them via wireless communication. However, as trajectory generation usually involves complex optimization problems, the maximum number of robots that can be handled is constrained by computational power. In contrast, in a distributed approach, each robot shares its planned trajectory with its neighbors via wireless communication, and the maximum number of robots that can be accommodated is limited by their communication capabilities. With trajectories available for all robots, each robot relies on a tracking controller to follow its respective trajectory. As planned trajectories contain higher-order information, differential flatness-based controllers are a suitable choice.

Due to their simplicity and ease of implementation, control-based methods are well-suited for managing large multi-robot systems, which are often referred to as ``robotic swarm" in the literature. These methods typically employ straightforward controllers capable of reacting promptly to obstacles or other robots, offering a fast and responsive response to a dynamic environment with low computation and communication resource requirements. Unlike multi-robot trajectory planning approaches, control-based methods directly govern robots' movements using velocity or acceleration commands based on global path and current local information. Popular examples of control-based methods include the artificial potential field method \cite{khatib1986real}, harmonic potential field method \cite{kim1992real}, navigation function method \cite{rimon1990exact}, and others. In recent years, the control barrier function method has also gained attention, taking the form of a quadratic programming problem with superior performance but requiring more significant computational resources. Importantly, some control-based approaches can operate autonomously without wireless communication and other robots' IDs, provided that robots are equipped with active detection devices such as cameras or radars.

Virtual tube planning and control is an innovative approach to navigating a multi-robot system through complex and cluttered environments. The virtual tube is just like the highway for vehicles, namely robots have no necessity to make collision avoidance maneuvers with obstacles in the environment, and they only need to keep inside the virtual tube. There is no obstacle within the virtual tube, and the area inside can be seen as a safety zone. Hence, there are three primary control objectives for robots inside the virtual tube, namely moving along the virtual tube, keeping within the virtual tube, and collision avoidance among robots. Other objectives can be considered according to the specific applications. There are many kinds of virtual tubes in the literature, such as straight-line virtual tube \cite{quan2021practical}, trapezoid virtual tube \cite{gao2022distributed,gao2022distributed2}, connected quadrangle virtual tube \cite{gao2022distributed}, annular virtual tube \cite{gao2022multi}, curve virtual tube \cite{quan2023distributed,gao2022robust}. Besides, some similar concepts have been proposed in the literature, such as the lane for autonomous road vehicles in \cite{rasekhipour2016potential}, \cite{luo2018porca} and the safe flight corridor for the multicopter in \cite{liu2017planning}. Additionally, the virtual tube can also be applied in air traffic management applications, in which virtual tubes form a complex ``sky highway'' \cite{quan2021sky,fu2022practical,safadi2023macroscopic}.

In this survey, we review all these methods in detail. We also conduct a comparative analysis of these methods and provide some subjective and general evaluations. We also explore in detail the various types of vehicle and sensor models, together with assumptions about obstacles and their movement.

\subsection{Exclusions}
In a multi-robot system, the software programs can be classified into three categories: perception, decision making, and multi-robot control. Besides, the simulation tools are also important. Due to the space limitation of this survey, we will not review all areas and only focus on the multi-robot passing through control. The following topics are exclusions of this survey, and we will only provide a brief summary where necessary.

\begin{itemize}
	\item \textit{Robot hardware system design.} The mechatronics design of each robot forms the bedrock of the multi-robot system. Several robots have been designed for various applications, and we classify them based on their kinematics in this survey. 
	\item \textit{High-level decision making.} High-level decision making has gained popularity in recent years. Multi-robot decision making can be categorized into two types: decision making for each robot and decision making for the whole multi-robot system. For passing-through control problems, the former involves generating goal points for every robot, whereas the latter produces a single goal point for the entire system. This survey presumes that all goal points are predetermined.
	\item \textit{Mapping methods.} Mapping is an important part of perception. As the multi-robot system is operated in a cluttered environment, the mapping methods to describe the environment are vital. Many collision avoidance techniques also require some kinds of digital map to operate. There also exist some local collision avoidance methods with no necessity to build a map. 
	\item \textit{Simulation tools.} Simulation is necessary to verify the effectiveness of the designed method. Simulations for the multi-robot system can be categorized into two types: numerical simulation and hardware-in-the-loop simulation. The former is mainly achieved by some programming languages and common software, such as MATLAB and Python. The implementation of the latter is based on some specific platforms, such as RflySim \cite{dai2021rflysim,wang2021rflysim} and AirSim \cite{shah2018airsim}.

\end{itemize}


\section{Problem Consideration}

In this section, we outline some of the realistic factors that influence the passing-through control problem. First, the possible control objectives are reviewed. Then, we review two multi-robot control schemes. At last, we review the modeling of the multi-robot system. 

\subsection{Control Objective for Passing-through Cluttered Environments}
There are many control objectives that can be considered in the passing-through controller. Besides the objective of all robots passing through the cluttered environment, two basic objectives related to the robots' safety are listed below:
\begin{itemize}
	\item All robots avoid collision with each other. 
	\item All robots  avoid collision with obstacles in the environment.
\end{itemize}
Besides, some other control objectives can also be considered for different situations.
\begin{itemize}
	\item Minimum passing-through time. In some literature, this objective is also called ``highest passing-through efficiency''. The cluttered environment is often dangerous for robots. Hence, the robots should pass through the cluttered environment as fast as possible.
	\item Minimum moving distance. This objective is similar to the one of the minimum passing-through time. A shorter moving distance usually corresponds to a higher passing-through efficiency.
	\item Minimum control effort. This objective may be necessary for robots operating in limited energy situations. For example, this objective is vital for multicopters with a limited flight time because of their battery capacity. This objective is also important for some fixed-wing UAVs and missiles with unpowered flying.
	\item Connectivity maintenance. In the passing-through process, this objective is usually used for maintaining the connectivity of the graph of communication or relative position. However, when the swarm controller is fully distributed, this objective may  not be necessary.
\end{itemize}

\subsection{Multi-Robot Control Scheme}
The primary considerations in multi-robot control revolve around information sharing and computational mechanisms. This subsection introduces two control schemes: centralized and distributed schemes. Furthermore, we outline the advantages and disadvantages associated with each scheme. In some literature, there exists another ``decentralized scheme''. In this survey, we classify the distributed scheme and the decentralized scheme into one category.

\subsubsection{Centralized Scheme}
In the centralized scheme, a central node is introduced. It can be a base station or a robot with strong computational ability in the multi-robot system. This central node monitors the whole multi-robot system to accomplish the passing-through task based on information gathered from all other robots. It is essential for all robots to maintain connectivity with the central node. The control algorithms usually have a good control performance with the centralized scheme. However, it has several drawbacks, including less robustness and heavy computational load. If the central node encounters a fault, the entire multi-robot system will fail. Additionally, the computational capabilities of individual robots are not fully utilized, and the communication resources face a burden due to the required connection links between the central node and other robots.

\subsubsection{Distributed Scheme}
In the distributed scheme, the multi-robot system has no need for the central node. The processing unit is available on the robot itself, and the control output is made by the robot based on its local observation. Compared to the centralized scheme, the distributed scheme is more suitable for a large robotic swarm. In some scenarios, the distributed scheme can operate autonomously without wireless communication and other robots' IDs. However, the implementation is much more difficult for a distributed system. Recently, the Defense Advanced Research Projects Agency (DARPA) put forward a novel concept called ``Mosaic Warfare''. The mosaic warfare places a premium on seeing battle as an emergent, complex system, and using low-cost unmanned swarms alongside other electronic and cyber effects to overwhelm adversaries. The central idea is to be distributed, cheap, fast, lethal, flexible, and scalable. In a word, the future direction of multi-robot control is towards the distributed scheme.

\subsection{Robot Kinematic Model}
There are many types of robots that have the necessity to operate in cluttered environments. All robots can be categorized into two types of kinematic models: the holonomic model and the non-holonomic model.

\subsubsection{Holonomic Model}
The holonomic model is applicable to robots with arbitrary orientation control capability. Namely, the holonomic model does not consider the robot's orientation. Examples of robots with holonomic models include multicopters, helicopters and certain types of wheeled robots equipped with omnidirectional wheels. 

Typical holonomic models include the single integrator model and the double integrator model. Assume that the multi-robot system composed of $M$ robots operates in a $n$-dimensional space. The single integrator model for the $i$th robot is shown as 
\begin{equation}
	\dot{\mathbf{{p}}}_{i} =\mathbf{v}_{\text{c},i}, \label{SingleIntegral}
\end{equation}
where $\mathbf{v}_{\text{c},i}\in {{\mathbb{R}}^{n}}$, $\mathbf{p}_{i}\in {{\mathbb{R}}^{n}}$ are the $i$th robot's velocity command and position, $i=1,\cdots ,M$. The single integrator model is usually viewed as an idealistic model for robots, as any robot can change its velocity instantaneously. The advantage of the single integrator model is that it can significantly simplify the analysis of the passing-through controller. Similarly, the double integrator model for the $i$th robot is shown as
\begin{align}
	\begin{cases}
		\dot{\mathbf{{p}}}_{i} =\mathbf{v}_{i} \\
		\dot{\mathbf{{v}}}_{i} =\mathbf{a}_{\text{c},i} 
	\end{cases},
	\label{DoubleIntegral}
\end{align}
in which $\mathbf{a}_{\text{c},i}\in {{\mathbb{R}}^{n}}$ indicates the acceleration command of the $i$th robot, and  $\mathbf{p}_{i},\mathbf{v}_{i}\in {{\mathbb{R}}^{n}}$ stand for the position and velocity of the $i$th robot, respectively. Because of its similarity to the well-known Newton's second law, the double integrator model is widely used in the literature. In some literature \cite{liu2020multi}, there also exist higher-order holonomic models, some of which are obtained from complex nonlinear systems after differential flatness.

In real practice, the velocity command is quite common for robots with holonomic kinematics. However, these robots cannot track their velocity commands as precisely as the single integrator model \eqref{SingleIntegral}. Here, a \emph{modified second-order model} can be used to describe this phenomenon \cite{quan2021practical}. The robot has a second-order holonomic model, and there exists a first-order process in the velocity loop to make the robot track its velocity command. The modified second-order model for the $i$th robot is shown as 
\begin{equation}
	\left\{
	\begin{aligned}
		\dot{\mathbf{{p}}}_{i} &=\mathbf{v}_{i}  \\
		\dot{\mathbf{{v}}}_{i} &=-l_i\left(\mathbf{v}_{i}-\mathbf{v}_{\text{c},i}\right)
	\end{aligned}
	\right., \label{Filtered}
\end{equation}
where $l_i>0$ relies on the physical properties and inner-loop controller of the $i$th robot. Then, a \emph{filtered position} can be defined to make a first-order model out of this second-order model. The filtered position of the $i$th robot is shown as \cite{quan2021practical}
\begin{equation}
	\boldsymbol{\xi}_{i}={\mathbf{p}}_{i}+\frac{1}{{l}_{i}}\mathbf{v}
	_{i}.  \label{FilteredPosition}
\end{equation}
Then it has $\boldsymbol{\dot{\xi}}_{i}  =\mathbf{v}_{\text{c},i}$, which is just like the single integrator model \eqref{SingleIntegral}.

\subsubsection{Non-holonomic Model}
Compared with holonomic models, non-holonomic models are suitable for robots with limited orientation control capability. Examples of robots with non-holonomic models include differential driving mobile robots, automated vehicles, fixed-wing UAVs, USVs, UUVs, etc.

There are many types of non-holonomic models, in which the representative ones are the unicycle model and the bicycle model. The unicycle model for the $i$th robot in $\mathbb{R}^2$ is shown as 
\begin{equation}
	\left\{
	\begin{aligned}
		\dot{x}_{i} &={v}_{\text{c},i}\cos\left(\psi_{i}\right) \\
		\dot{y}_{i} &={v}_{\text{c},i}\sin\left(\psi_{i}\right) \\
		\dot{\psi}_{i} &=\omega_{\text{c},i}
	\end{aligned}
	\right.,\label{Unicycle}
\end{equation}
where $\mathbf{p}_{i}=\left[{x}_{i}\ y_{i}\right]^{\text{T}}$ and $\mathbf{v}_{i}=\left[\dot{x}_{i}\ \dot{y}_{i}\right]^{\text{T}}$ are the position and velocity of the $i$th robot, $\psi_{i}$ is the heading (yaw) angle, and ${v}_{\text{c},i}$ and $\omega_{\text{c},i}$ are the speed command and the yaw angular speed command. 
Different from the unicycle model, the bicycle model simplifies the robot to a vehicle, which consists of a steerable front wheel and a fixed rear wheel. The bicycle model for the $i$th robot in a two-dimensional plane is shown as \cite{hoy2015algorithms}
\begin{equation}
	\left\{
	\begin{aligned}
		\dot{x}_{i} &={v}_{i}\cos\left(\psi_{i}+\beta\left(\omega_{\text{c},i}\right)\right) \\
		\dot{y}_{i} &={v}_{i}\sin\left(\psi_{i}+\beta\left(\omega_{\text{c},i}\right)\right) \\
		\dot{v}_{i} &={a}_{\text{c},i}\\
		\dot{\psi}_{i} &= \frac{{v}_{i}}{{l}_{\text{r}}}\sin\left(\beta\left(\omega_{\text{c},i}\right)\right)
	\end{aligned}
	\right.,\label{BicycleModel}
\end{equation}
where $\mathbf{p}_{i}=\left[{x}_{i}\ y_{i}\right]^{\text{T}}$ and $\mathbf{v}_{i}=\left[\dot{x}_{i}\ \dot{y}_{i}\right]^{\text{T}}$ are the position and velocity of the $i$th robot, $\psi_{i}$ is the yaw angle, ${l}_{\text{r}}$ is the distance from the rear wheel to the center of gravity, ${a}_{\text{c},i}$ and $\omega_{\text{c},i}$ are the acceleration command and the yaw angular speed command. The variable $\beta\left(\omega_{\text{c},i}\right)$ represents the sideslip angle of the center of gravity, which is calculated as 
 \begin{equation*}
\beta\left(\omega_{\text{c},i}\right)=\arctan\left(\tan\left(\omega_{\text{c},i}\right)\frac{{l}_{\text{r}}}{{l}_{\text{f}}+{l}_{\text{r}}}\right),
 \end{equation*}
where ${l}_{\text{f}}$ is the distance from the front wheel to the center of gravity.

\subsection{Robot Geometric Model}
In the literature, five kinds of areas around a robot are typically considered, which are \emph{physical area}, \emph{safety area}, \emph{avoidance area}, \emph{communication area} and \emph{detection area}. In the following, these areas are introduced in a two-dimensional plane.

\subsubsection{Physical Area}
The physical area is used to represent the robot's space occupied in the environment. A circular physical area $\mathcal{P}_{i}$ of the $i$th robot in $\mathbb{R}^2$ is defined as
\begin{equation*}
	\mathcal{P}_{i}=\left \{ \mathbf{x}\in {{\mathbb{R}}^{2}}: \left \Vert \mathbf{x}-\mathbf{p}_{i}\right \Vert \leq r_{\text{p}} \right \} ,
\end{equation*}
where $r_{\text{p}}>0$ is the physical radius. For all robots, no conflict with each other implies that $\mathcal{P}_{i}\cap \mathcal{P}_{j}=\varnothing$,
namely $\left \Vert \mathbf{p}_{i}-\mathbf{p}_{j}\right \Vert >2r_{\text{p}}$,
where $i,j=1,\cdots ,M,i\neq j$.

\subsubsection{Safety Area}
The safety area is usually related to the control objective of collision avoidance. A circular safety area $\mathcal{S}_{i}$ of the $i$th robot in $\mathbb{R}^2$ is defined as
\begin{equation*}
	\mathcal{S}_{i}=\left \{ \mathbf{x}\in {{\mathbb{R}}^{2}}: \left \Vert \mathbf{x}-\mathbf{p}_{i}\right \Vert \leq r_{\text{s}} \right \} ,
\end{equation*}
where $r_{\text{s}}>0$ is the safety radius. We have $r_{\text{s}}>r_{\text{p}}$. Due to the occurrence of unexpected uncertainties, just ensuring $\mathcal{P}_{i}\cap \mathcal{P}_{j}=\varnothing$ for all robots avoiding collision with each other may be insufficient. Hence, the control objective is usually set as  $\mathcal{S}_{i}\cap \mathcal{S}_{j}=\varnothing$, where $i,j=1,\cdots ,M,i\neq j$.

\subsubsection{Avoidance Area}
The avoidance area is defined for starting the avoidance control. A circular avoidance area $\mathcal{A}_{i}$ of the $i$th robot in $\mathbb{R}^2$ is defined as 
\begin{equation*}
	\mathcal{A}_{i}=\left \{ \mathbf{x}\in {{\mathbb{R}}^{2}}:\left
	\Vert \mathbf{x}- \mathbf{p}_{i}\right \Vert \leq r_{\text{a}}
	\right \}  ,
\end{equation*}
where $r_{\text{a}}>0$ is the avoidance radius. We have $r_{\text{a}}>r_{\text{s}}>r_{\text{p}}$.

\subsubsection{Communication Area}
The communication area is used to represent the robot's communication capacity. A circular communication area $\mathcal{C}_{i}$ of the $i$th robot in $\mathbb{R}^2$ is defined as 
\begin{equation*}
	\mathcal{C}_{i}=\left \{ \mathbf{x}\in {{\mathbb{R}}^{2}}:\left
	\Vert \mathbf{x}- \mathbf{p}_{i}\right \Vert \leq r_{\text{c}}
	\right \}  ,
\end{equation*}
where $r_{\text{c}}>0$ is the avoidance radius. We have $r_{\text{c}}>r_{\text{a}}>r_{\text{s}}>r_{\text{p}}>0$. When the $j$th robot is inside $\mathcal{C}_{i}$, namely $\mathbf{p}_j \in \mathcal{C}_{i}$, the $i$th robot can communicate with the $j$th robot. This communication link can be unidirectional or bidirectional. If it is unidirectional, the communication graph is undirected. Otherwise, the communication graph is a directed graph.

\subsubsection{Detection Area}
\begin{figure}[!t]
	\begin{centering}
		\includegraphics[scale=1]{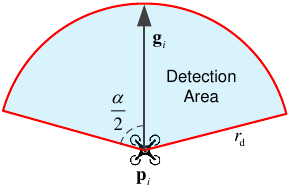}
		\par \end{centering}
	\caption{Fan-shaped detection area of a robot.}
	\label{DetectionArea}
\end{figure}

The detection area is used to represent the robot's detection capacity. In many cases, each robot in the multi-robot system is equipped with an active detecting device, like a camera or a radar, to detect the relative positions of other robots. The orientation angle of the active detecting device of the $i$th robot is represented as ${\eta}_{i}\in \left[-\pi,\pi\right]$. The vector $\mathbf{g}_i=\left[ \cos{\eta}_{i}\ \sin{\eta}_{i}\right]^{\text{T}}$ depicts the unit gazing vector of the $i$th robot. As shown in Figure \ref{DetectionArea}, a fan-shaped detection area $\mathcal{D}_{i}$ of the $i$th robot in $\mathbb{R}^2$ is defined as 
\begin{align*}
	\mathcal{D}_{i}=&\left \{ \mathbf{x}\in {{\mathbb{R}}^{2}}: \left \Vert \mathbf{x}-\mathbf{p}_{i}\right \Vert \leq r_{\text{d}},\left\langle \mathbf{g}_i,\mathbf{x}-\mathbf{p}_{i}\right\rangle <\frac{\alpha}{2} \right \} ,
\end{align*}
where $r_{\text{d}}>0$ is the detection radius, $0<\alpha\leq2\pi$ is the view field angle, $\left\langle \mathbf{a},\mathbf{b}\right\rangle$ represents the angle between two vectors $\mathbf{a}$ and $\mathbf{b}$. When the $j$th robot is inside $\mathcal{D}_{i}$, namely $\mathbf{p}_j \in \mathcal{D}_{i}$, the $i$th robot can detect the $j$th robot.

\section{Leader-Follower Formation Control}

When robots are passing through a cluttered environment, the sensed variables and the corresponding controlled variables are crucial for a leader-follower formation. The type of sensed variables determines the requirements for the perception capabilities of robots. The type of controlled variables is highly related to the topology of the wireless communication network. In this section, we categorize some existing results on leader-follower formation control into displacement-based control, distance-based control and bearing-based control. In some literature, the position-based control can also generate a formation. However, such methods are usually unsuitable for the passing-through task. As shown in Figure \ref{Compare}, the main advantage of the leader-follower formation control is its theoretical completeness.

\subsection{Displacement-based Leader-Follower Formation Control}
With displacement-based formation control, robots actively control displacements of their neighboring robots to achieve a desired formation. It is assumed that each robot is able to sense the relative positions of its adjacent robots with respect to the global coordinate system. This implies that each robot needs to know the orientation of the global coordinate system but has no need to know its position in the global coordinate system. In displacement-based control, the requirement for the network topology is that the entire network should be either a connected undirected graph or a directed graph with at least one directed spanning tree. If the network topology is time-varying, the entire network needs to have consistent topological connectivity.

\begin{figure}[!t]
	\begin{centering}
		\includegraphics[width=\columnwidth]{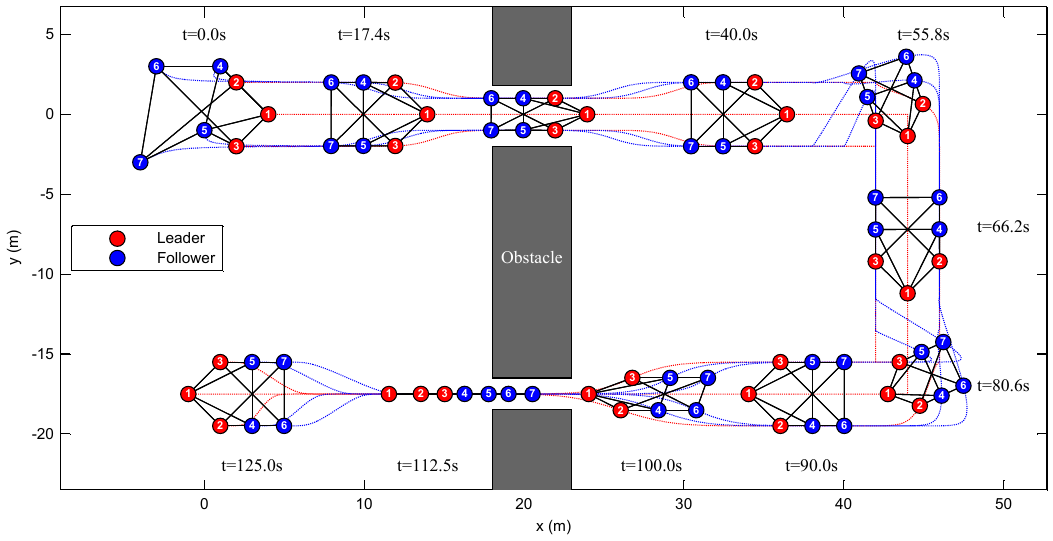}
		\par \end{centering}
	\caption{A simulation of affine formation maneuver control \cite{zhao2018affine}. }
	\label{AffineFormation}
\end{figure}

In recent years, several works have been proposed to solve the displacement-based formation control problem. In \cite{hu2014adaptive}, the authors propose a second-order distributed adaptive formation tracking algorithm based on the leader-follower formation framework. With this algorithm, each robot can achieve formation control by utilizing only the relative position and relative velocity information of its neighboring robots. As shown in Figure \ref{AffineFormation}, the authors in \cite{zhao2018affine} introduce an affine formation maneuvering control method to realize the formation of geometric shapes in a multi-agent system. This method allows for continuous and smooth adjustments of the formation's center, orientation, and size. Building upon this, the authors in \cite{chen2020distributed} further investigate affine formation maneuvering control in high-order multi-agent systems. The agents are divided into three types: the first leader, the second leaders, and the followers. Addressing the control problem in heterogeneous multi-agent systems, the authors in \cite{xu2019affine} propose a proportional-integral control scheme to ensure the formation structure based on the leader-follower approach and successfully eliminate steady-state errors.

\subsection{Distance-based Leader-Follower Formation Control}
With distance-based formation control, each robot gets its distance to other adjacent robots through perception devices, such as a camera and a ultra wide band (UWB) device. Robots actively control their movements to approach the desired distances with other robots, thereby achieving the desired formation structure. The entire formation can be treated as a rigid body, indicating that the network topology requires a rigid structure. In distance-based formation, each robot needs to have its own local coordinate system. Unlike displacement-based formations, the local coordinate systems have no need to be aligned with each other.

In \cite{anderson2008rigid}, the authors investigate the minimum rigidity theory, which states that removing any edge from a wireless communication topology will cause the entire network to lose its rigidity structure. The minimum rigidity theory provides important support for studying distance-based formation control algorithms. In \cite{krick2009stabilisation}, the authors propose a gradient vector field control algorithm to achieve a distance-based formation structure. In \cite{rozenheck2015proportional}, the authors present a distance-based formation control algorithm following a leader-follower structure. The leader tracks its desired velocity command, while the followers maintain their desired distances and follow the leader, thereby achieving the desired formation structure. In \cite{bae2020distributed}, the authors consider external disturbances and propose a distributed robust adaptive gradient controller for controlling distance-based formation. Based on the Lyapunov stability analysis of the minimum infinite rigidity formation system, this paper obtains an upper bound for the square of distance errors and verifies the effectiveness of the theoretical results through numerical simulations.

\subsection{Bearing-based Leader-Follower Formation Control}

With bearing-based formation control, each robot perceives its bearing to other adjacent robots. They actively control their movements to align themselves with the desired bearings to other robots, thereby achieving the desired formation structure. The network topology requires a bearing rigidity structure. In bearing-based formation, each robot needs to have its own local coordinate system, and the local coordinate systems need to be aligned with each other. 

In \cite{zhao2019bearing}, the authors provide a comprehensive review of the bearing rigidity theory and its applications. The bearing rigidity theory serves as the fundamental theory for supporting research on bearing-based formation control. In \cite{zhao2015bearing}, a distributed formation control algorithm is designed based on the bearing rigidity theory, allowing for scalable formation. The scaling of the formation is uniquely determined by a translation factor and a scaling factor, ensuring a stable formation structure. In \cite{li2020adaptive}, the authors investigate the formation control problem in three-dimensional space with parameter uncertainties, where each robot can only perceive the bearing information from its neighbors. This paper extends the bearing rigidity theory from point-mass models to Euler-Lagrange nonlinear models and proposes an almost globally stable formation control algorithm using the backstepping control method.

\begin{figure*}[!t]
	\begin{centering}
		\includegraphics[width=\textwidth]{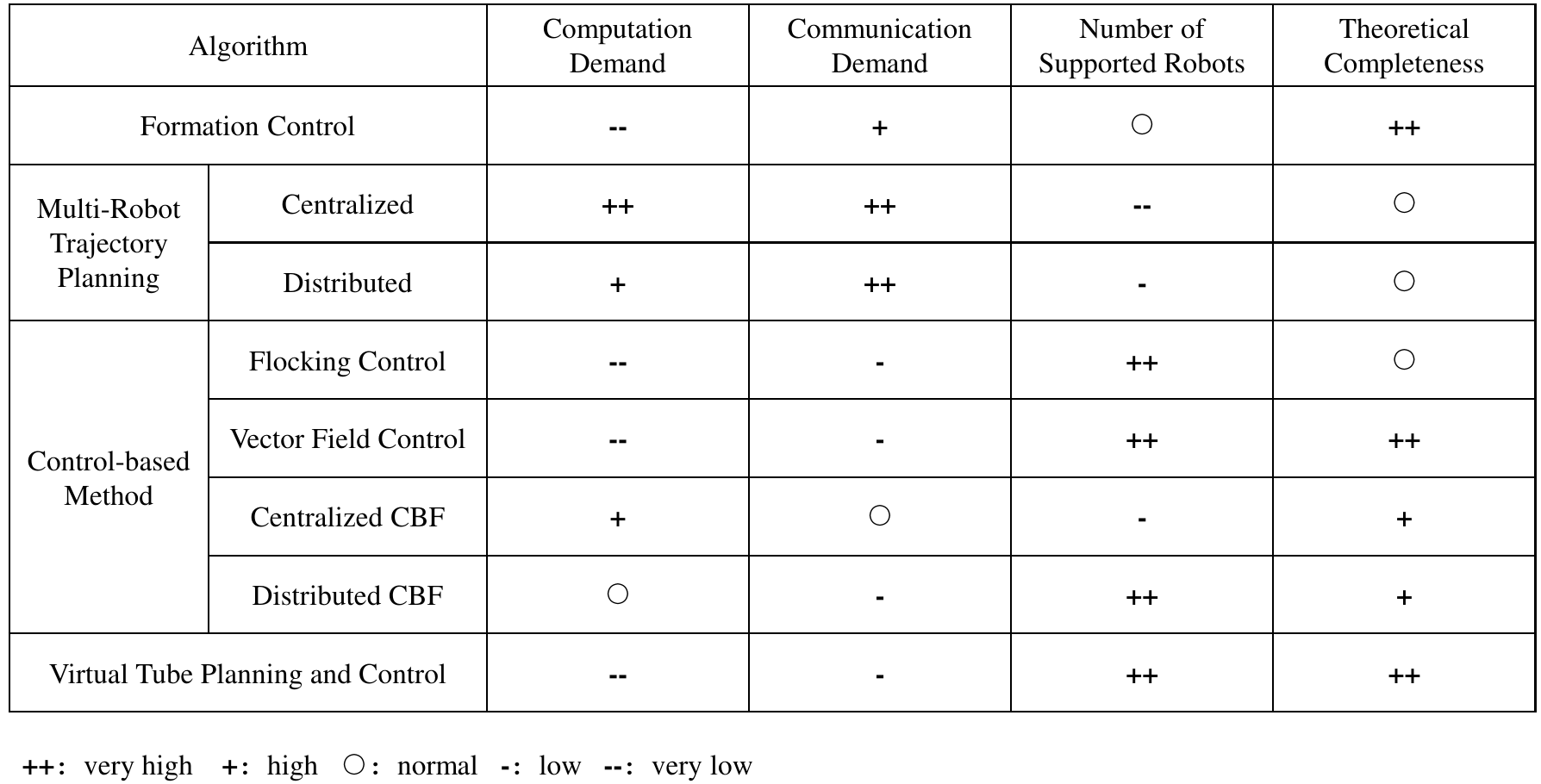}
		\par \end{centering}
	\caption{A comparison of algorithms for passing through cluttered environments.}
	\label{Compare}
\end{figure*}

\section{Multi-Robot Trajectory Planning}
The foundation of multi-robot trajectory planning lies in the trajectory planning of a single robot. In this section, we first review some methods related to the trajectory planning of a single robot. Then, the methods related to multi-robot path planning are reviewed. When paths for robots are available, centralized or distributed trajectory planning methods can be used to generate smooth trajectories for all robots. As shown in Figure \ref{Compare}, the main advantage of multi-robot trajectory planning is its wonderful control performance.

\subsection{Single-Robot Trajectory Planning}
In the task of a single robot passing through a cluttered environment, it is common to construct a hierarchical trajectory planning framework. This involves first searching for a discrete path in a known global map and then optimizing the path to form a smooth and continuous trajectory in the local map. In real practice, the known global map can be constructed from pre-installed information such as satellite images or building blueprints. Obviously, such rough maps are insufficient to support robots in completing tasks in cluttered environments. Therefore, robots need to continuously supplement and improve the digital map during the passing-through process by relying on their own sensing devices.

\subsubsection{Single-Robot Path Planning}
In the single-robot hierarchical trajectory planning framework, the first step is to search for a discrete geometric path in a known digital map based on the coordinates of the target point. Common path planning algorithms can be divided into two categories: search-based path planning algorithms and sampling-based path planning algorithms. Search-based path planning algorithms include Dijkstra algorithm \cite{dijkstra2022note}, A* algorithm \cite{hart1968formal}, ARA* algorithm \cite{likhachev2003ara}, Hybrid A* algorithm \cite{dolgov2008practical}, JPS algorithm \cite{harabor2011online}. The sampling-based path planning algorithms include the PRM algorithm \cite{geraerts2004comparative}, RRT algorithm \cite{lavalle2001randomized}, RRT* algorithm \cite{karaman2010incremental}. Some detailed reviews of these path planning algorithms can be found in \cite{patle2019review,aggarwal2020path,cheng2021path}.

\subsubsection{Single-Robot Trajectory Planning}
After the path planning, the second step is to optimize the discrete path into a smooth and continuous trajectory. The commonly used methods for describing trajectories include cubic spline interpolation curves \cite{chen2020dynamic}, Bézier curves \cite{zhou2020robust}, B-spline curves \cite{askari2016new}, MINCO curves \cite{wang2022geometrically}, etc. Many trajectory planning algorithms have been proposed. The authors in \cite{mellinger2012mixed} propose a mixed integer quadratic programming (MIQP) optimization method to solve the obstacle avoidance problem in complex environments. Similarly, the authors in \cite{culligan2006online} propose a mixed integer linear programming (MILP) optimization method. However, both the MIQP and MILP algorithms require a high computation demand, which makes them only suitable for offline trajectory planning in known environments.

\begin{figure}[!t]
 	\begin{centering}
 		\includegraphics[width=\columnwidth]{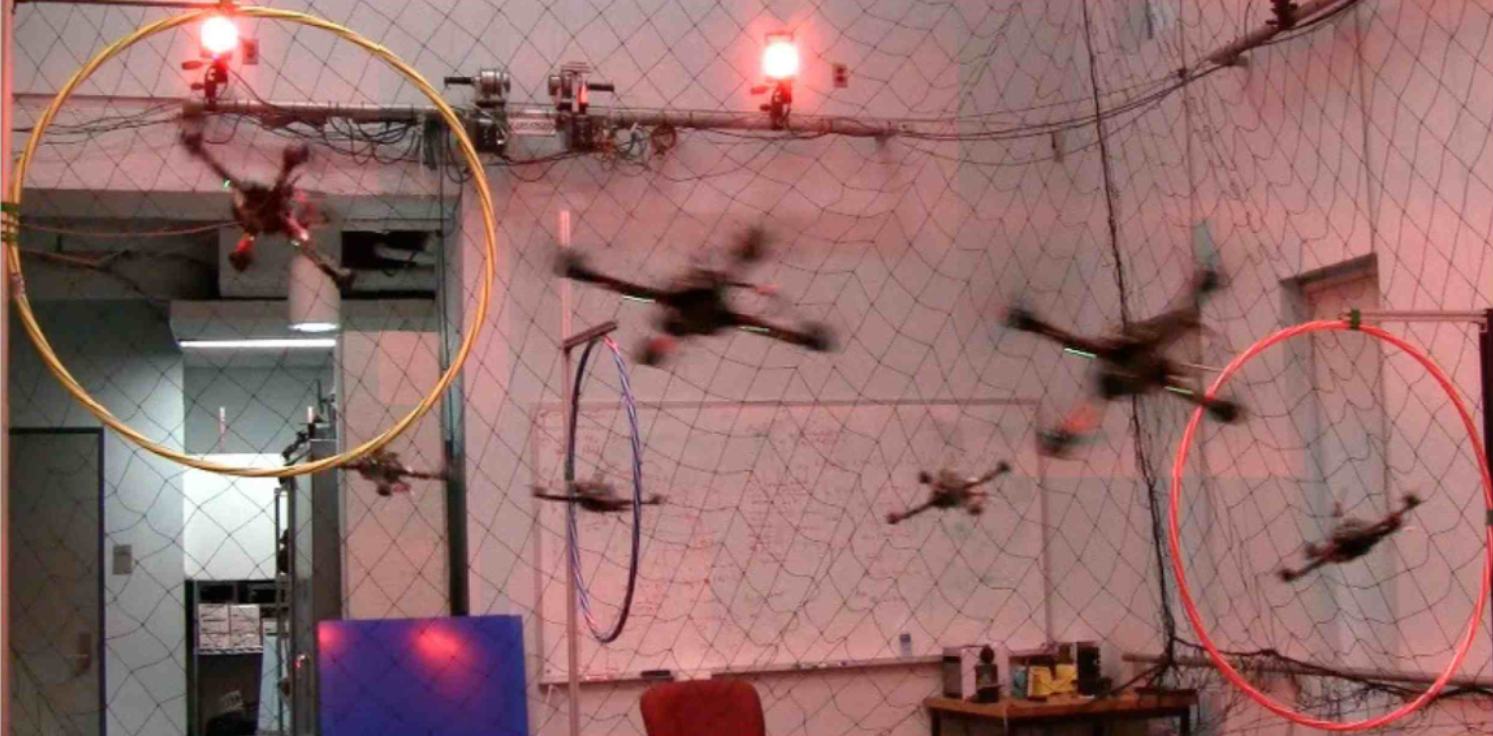}
 		\par \end{centering}
 	\caption{Experiment based on the minimum snap algorithm \cite{mellinger2011minimum}. }
 	\label{MinimumSnap}
\end{figure}

In \cite{van1998real}, the authors solve the trajectory planning problem in two steps. Firstly, a smooth trajectory is generated based on multiple constraints. Secondly, a nonlinear controller based on the differential flatness is used to track the trajectory. In \cite{mellinger2011minimum}, the authors develop a multicopter trajectory planning framework ``Minimum Snap'' as shown in Figure \ref{MinimumSnap}. In \cite{richter2016polynomial}, the authors improve the minimum snap algorithm by converting it into an unconstrained optimization problem and obtaining a closed-form solution. In \cite{chen2016online}, the authors propose a method for multicopters to generate real-time collision-free trajectories, which can quickly respond to new obstacles detected by sensors and replan the flight trajectory to ensure flight safety.
 
\subsection{Multi-Robot Path Planning}

Regarding the task of passing through cluttered environments, the algorithms commonly used for single-robot path planning are also utilized in multi-robot path planning. To avoid collision among robots, these paths should have as few intersections as possible.

\subsubsection{Search-based Multi-Robot Path Planning Algorithm}
In \cite{mac2017hierarchical}, the Dijkstra algorithm is used to obtain geometric paths for each robot in the digital map without collision through triangulation. In \cite{bai2019distributed}, the authors propose a multi-car path planning algorithm based on the Dijkstra algorithm, which minimizes the travel time between any two given points. However, the algorithm proposed in this paper can only be used in static environments and cannot be applied to dynamic scenes. In \cite{erokhin2019optimal}, an improved A* algorithm is proposed for coordinated path planning of a robotic swarm. This algorithm assigns dynamic values to each node, such that when a robot's path passes through a node, its assigned value changes dynamically. Other robots will try to avoid this node as much as possible and generate their own paths. In \cite{sun2019novel}, the A* algorithm is combined with other algorithms to solve the problem of large-scale coverage of robot swarms. In this paper, the A* algorithm is not used to generate paths, but rather to optimize the nominal paths to minimize the distance traveled by each robot.

\subsubsection{Sampling-based Multi-Robot Path Planning Algorithm}
In \cite{preiss2017downwash}, a trajectory planning method is proposed for large-scale multicopter UAV swarms, in which the PRM algorithm considering the downwash of UAVs is used in the path planning stage. In \cite{madridano20203d}, the authors use the PRM algorithm to generate non-conflicting flight paths in a digital map. A major advantage of the PRM algorithm for cooperative path planning is its ability to efficiently perform planning in three-dimensional space. In \cite{wu2020probabilistically}, the authors modify the RRT* algorithm to propose a multi-UAV path planning algorithm for urban air traffic, which can create collision-free trajectories under the influence of external disturbances. In \cite{berning2020rapid}, the authors use the dynamic RRT* algorithm to plan the flight paths of heterogeneous UAVs in urban environments. This paper verifies the feasibility of these flight paths through linear covariance propagation and collision detection algorithms based on quadratic programming.

\subsection{Multi-Robot Trajectory Planning}
Similar to the trajectory planning for a single robot, once the discrete geometric paths of each robot are obtained, the next step is to convert these paths into smooth trajectories. Multi-robot trajectory planning algorithms can be divided into two categories: centralized and distributed. Centralized algorithms have a central computing node that calculates trajectories for all robots, and such architecture cannot meet the computational demands of large-scale swarms. Therefore, multi-robot distributed trajectory planning algorithms have a wider range of applications.

\subsubsection{Multi-Robot Centralized Trajectory Planning}

Based on the MILP trajectory planning algorithm for a single robot, the authors in \cite{song2016rolling} propose a MILP cooperative trajectory planning algorithm suitable for UAV swarms, which consider some constraints such as the initial position of the UAV, the target point location, and battery power. 
In \cite{lal2017optimal}, the authors design a MILP trajectory planning algorithm for UAV swarms to perform pesticide spraying tasks. To ensure the completeness of spraying, UAVs are required to visit all known nodes. 
Based on the single robot MIQP trajectory planning algorithm, the authors in \cite{mellinger2012mixed} introduce an algorithm that uses the MIQP algorithm to generate three-dimensional smooth trajectories for heterogeneous multicopter swarms in complex environments full of obstacles. This algorithm pays special attention to the smoothness at the connection point between two trajectories. In order to reduce the computational complexity of the MIQP algorithm, the authors in \cite{kushleyev2013towards} divide the entire UAV swarm into multiple rigid formations, and plan trajectories for these formations, which reduce the computational complexity. In summary, although MILP and MIQP algorithms can obtain ideal optimization results, their computational complexity is usually very large and they are not suitable for real-time trajectory planning for robotic swarms.

The model predictive control (MPC) algorithm follows the rolling horizon optimization concept and can also be used in centralized trajectory planning for robotic swarms. In \cite{soria2021predictive}, a nonlinear MPC algorithm is proposed to solve the navigation problem of multicopter swarms in complex environments. In \cite{zhang2019trajectory}, the trajectory planner and tracking controller of autonomous ground vehicles are integrated to generate and track smooth trajectories, while achieving collision avoidance with other vehicles and obstacles. The trajectory planner is designed based on a state machine, and the tracking controller is designed based on the MPC using the vehicle's kinematic model.

\subsubsection{Multi-Robot Distributed Trajectory Planning}

\begin{figure}[!t]
	\begin{centering}
		\includegraphics[scale=0.3]{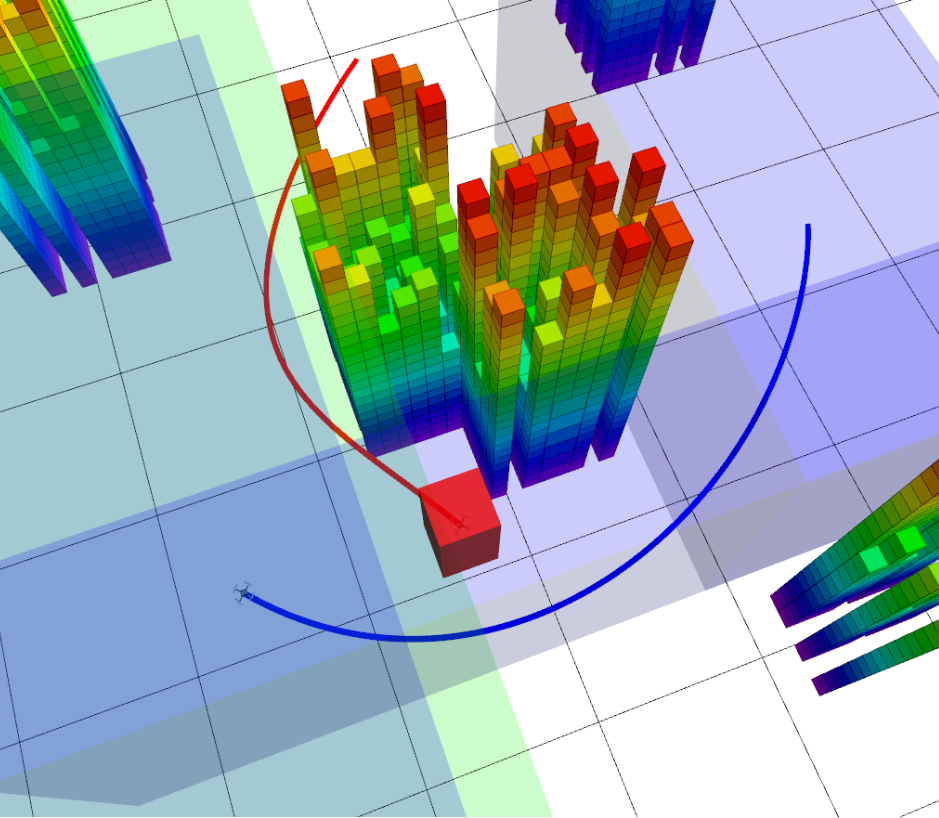}
		\par \end{centering}
	\caption{Trajectory planning for multiple robots using safe flight corridor and relative safe flight corridor \cite{park2019fast}. }
	\label{FlightCorridor}
\end{figure}

In order to apply the concept of safe flight corridor in a distributed manner, the authors in \cite{park2020online,park2019fast} replace the time-varying and non-convex collision avoidance constraints of the relative safe flight corridor with convex constraints based on the idea of velocity obstacle. As shown in Figure \ref{FlightCorridor}, these papers successfully achieve distributed collision-free trajectory planning for multicopter swarms, while reducing the total flight time and distance without sacrificing the success rate. 

In \cite{park2022online}, the authors further improve the relative safe corridor to a linear safe corridor, which does not rely on slack variables or soft constraints to avoid optimization failure. In addition, this paper uses a priority-based goal programming approach to prevent the deadlock problem in distributed optimization, without requiring additional communication for swarm decision-making.
In \cite{luis2020online}, the authors propose a distributed MPC method for generating collision-free trajectories for a robotic swarm. Additionally, the paper introduces an event-triggered re-planning strategy to cope with external disturbances. 
In \cite{zhou2021ego,zhou2021decentralized}, the authors propose a distributed trajectory planning system for a multicopter swarm relying solely on on-board sensing devices. This system is established based on a gradient-based local trajectory planning framework, where collision risks among UAVs are introduced as a penalty term in the nonlinear optimization problem.

\section{Control-based Methods}
With control-based methods, each robot in the robotic swarm uses the same or similar controllers to form a swarm behavior. Unlike the formation control, robots do not need to maintain a fixed geometric structure. In this section, we will introduce several control-based methods for robotic swarms, including flocking control, vector field control, and control barrier function. 

\subsection{Flocking Control}
In the literature, the flocking is considered a form of collective behavior that is caused by individuals following simple rules without any central coordination. As shown in Figure \ref{Compare}, the main advantage of the flocking control is that it is suitable for a large robotic swarm.  
The concept of the flocking control originates from the well-known Boids model \cite{reynolds1987flocks,liu2023task}. The Boids model involves three flocking behaviors, which are introduced as follows.
\begin{itemize}
	\item Collision avoidance behavior. This behavior is used to avoid collision with adjacent robots.
	\item Cohesion behavior. This behavior is used to guide the robot to the average position of adjacent robots.
	\item Velocity alignment behavior. This behavior is used to make the robot achieve speed synchronization with adjacent robots.
\end{itemize}
It should be noted that these three behaviors are all local rules, and there is no ``leader" playing a role in the robotic swarm. Any robot only adjusts its own speed and direction based on the movement of other adjacent robots.

On the basis of the Boids model, Viscek simplifies it and studies the collective behavior of particles from the perspective of statistical mechanics, and proposes the famous Viscek model \cite{vicsek1995novel}. In this model, each particle can achieve its own velocity parallel to that of its surrounding neighbor particles, but the motion of each particle is also affected by noises. By adjusting the noise and density, it is found that the particles as a whole can exhibit a transition between random motion and collective motion. Besides, some studies have made further improvements to simulate the collective motion of groups more accurately, such as the Couzin model \cite{couzin2002collective} and the Cucker-Smale model \cite{cucker2007emergent}.


The aforementioned flocking control methods can be used to guide the robotic swarm in a cluttered environment. In \cite{la2011flocking}, the authors propose an improved bio-inspired flocking algorithm for controlling a double-integrator multi-agent system to navigate through complex environments filled with obstacles. In \cite{ban2021self}, the authors use a modified flocking algorithm to control a heterogeneous robotic swarm to complete various complex tasks while ensuring that the robots do not collide with each other. The entire robotic swarm designates a leader to communicate target information to the entire swarm.

To improve the efficiency of the flocking control, the optimal flocking algorithm has become a recent research hotspot. In \cite{beaver2020optimal}, the authors utilize a distributed optimal flocking algorithm to guide the motion of a multi-agent system, which can achieve real-time computation while ensuring safety and energy optimization. In \cite{beaver2020beyond}, the authors propose a bio-inspired optimal flocking algorithm by transforming the rules of the Boids model into an optimal control problem. This paper also demonstrates that when the speeds of all agents are consistent, the multi-agent system reaches global optimality. In \cite{liu2021hierarchical}, the authors improve the Viscek model and design a hierarchical optimal flocking algorithm for multicopter swarms. All multicopters are divided into three layers with different weights, which can achieve faster alignment convergence and more flexible collision avoidance among multicopters.

\subsection{Vector Field Control}

\begin{figure}[!t]
	\begin{centering}
		\includegraphics[scale=0.4]{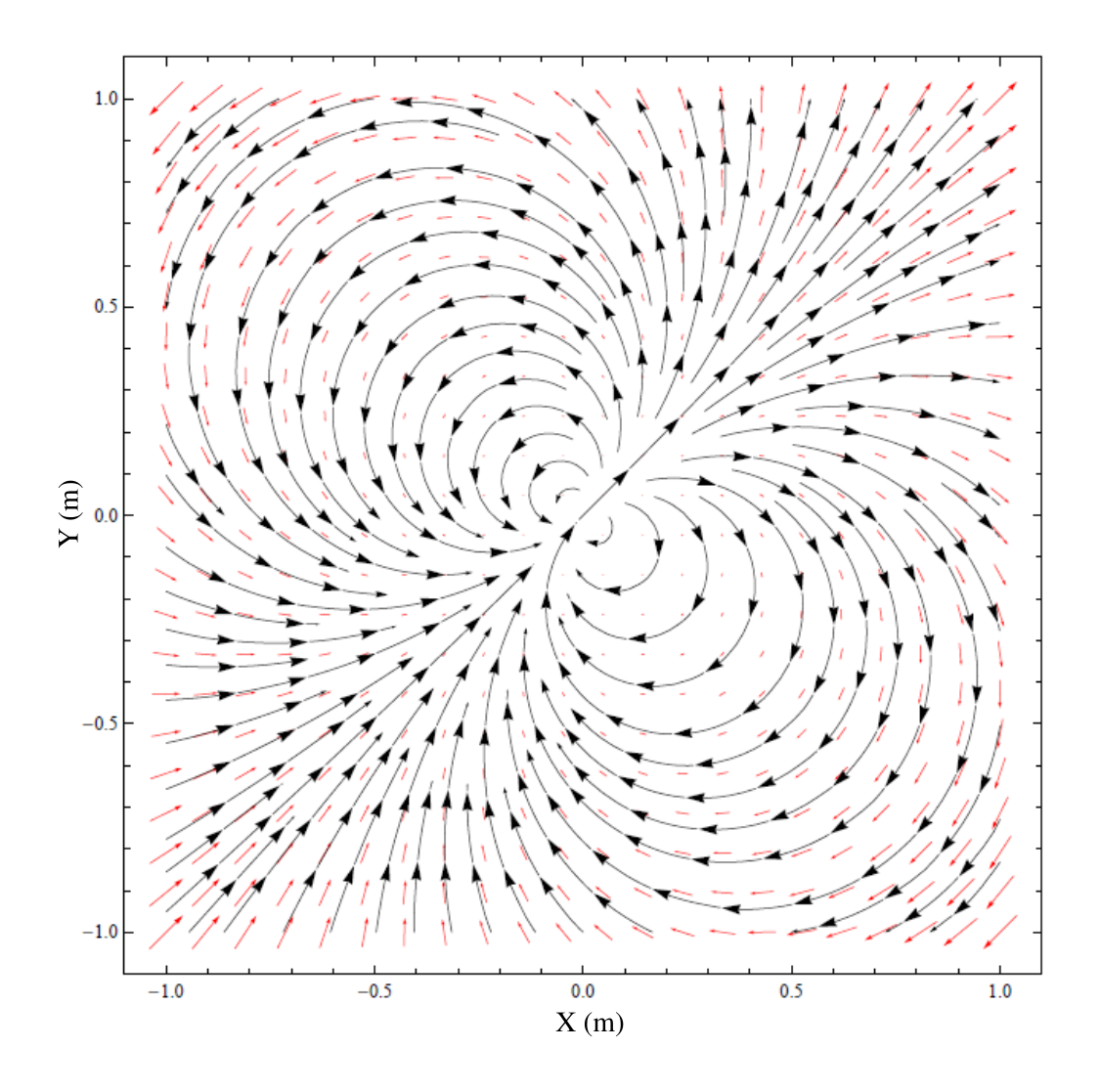}
		\par \end{centering}
	\caption{A typical non-gradient vector field \cite{panagou2014motion}.}
	\label{Nongradient}
\end{figure}

The vector field algorithms are widely used for the distributed control of robotic swarms. As shown in Figure \ref{Compare}, the main advantage of the vector field algorithm is its theoretical completeness, and it is suitable for a large robotic swarm. The vector field can be classified into two types: gradient vector field and non-gradient vector field. According to the description in \cite{arfken1999mathematical}, a vector field is a gradient vector field if it is the gradient of a scalar potential function. The curl of a gradient vector field is zero, which means that it is irrotational. Typically, the design of a gradient vector field algorithm involves first designing a potential function, and then obtaining the vector field of the robot directly through gradient operations. In contrast, the non-gradient vector field algorithm refers to the vector field that cannot find the corresponding potential function. Figure \ref{Nongradient} presents a typical non-gradient vector field \cite{panagou2014motion}. In the following, we will first introduce the gradient vector field algorithm, which includes the artificial potential field method and the Lyapunov guidance vector field method, and then continue to introduce the non-gradient vector field algorithm.

\subsubsection{Artificial Potential Field Method}

The most widely used algorithm for gradient vector fields is the artificial potential field (APF) method. The APF method was first proposed by Khatib in 1985 \cite{khatib1986real}.  Compared to the control barrier function method introduced in the following, the APF method is particularly suitable for handling multi-objective control problems. Each control objective can be designed as an attractive or repulsive potential field function. For example, the goal point exerts ``attraction" on the robot, while obstacles or other robots exert ``repulsion" on the robot. By taking the gradient of all potential field functions, the robot's velocity command can be obtained. However, the APF method has several imperfections, with the most serious problem being the local minima problem, which is also known as the deadlock problem \cite{hernandez2011convergence}. The local minima problem arises when unexpected equilibrium points occur where the potential field is equal to zero, causing the robot to become trapped in these local minimum points and unable to reach the goal point.

\begin{figure}[!t]
	\begin{centering}
		\includegraphics[width=\columnwidth]{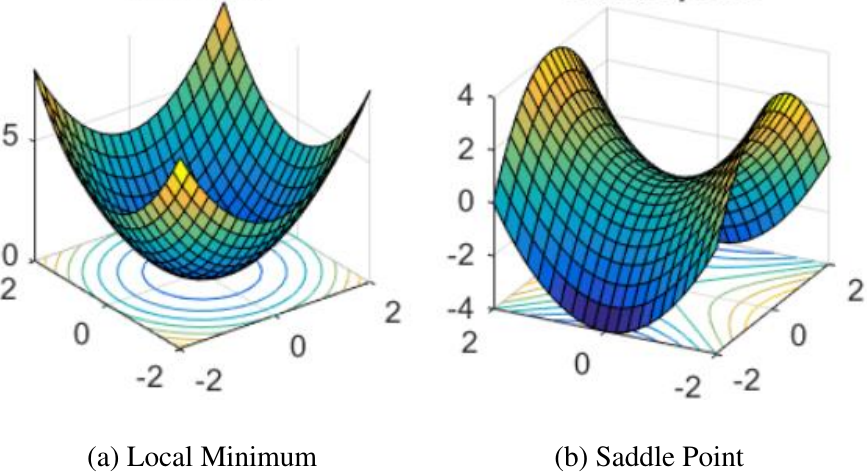}
		\par \end{centering}
	\caption{The comparison of local minimum and saddle point.}
	\label{Saddle}
\end{figure}

Currently, there are three main approaches dedicated to solving the local minima problem. The first approach is to keep the robot away from the local minimum or force it to leave the local minimum. The authors in \cite{rostami2019obstacle} decompose the repulsive vector field into two components, which are parallel and perpendicular to the attractive vector field. The authors only retain the component perpendicular to the attractive vector field. In \cite{antich2005extending,ge2005queues}, the authors first detect whether the robot has fallen into a local minimum point, and if it has, they apply extra forces to help it escape. The second approach combines the APF method with optimization algorithms to eliminate local minima. The authors in \cite{vadakkepat2000evolutionary} combine the APF method with a genetic algorithm to obtain an optimal APF method, which is called evolutionary APF. The authors in \cite{orozco2019mobile} further improve the evolutionary APF method into membrane evolutionary APF method, which can achieve better results in a shorter time. The third approach is to directly study the form of the potential field function to ensure that all vector field singularities are saddle points rather than local minimum points. Saddle points can be regarded as isolated points with the Lebesgue measure equal to zero, and the robot will not stay at the saddle point. If the Hessian matrix of the vector field at a singularity point is nonsingular, then the point is a saddle point. The comparison of local minimum and saddle point is presented in Figure \ref{Saddle}. To make all vector field singularities become saddle points, the authors in \cite{kim1992real} propose a harmonic potential field function, which can prevent the robot from falling into local minimum points. In addition, there are algorithms whose potential functions are not obtained by linear superposition. Among them, the most representative one is the navigation function method \cite{rimon1990exact}, which can be considered as a variation of the APF method, but its potential function takes the form of Morse function \cite{koditschek1990robot}. When parameters in the Morse function satisfy certain conditions, it is possible to achieve the absence of local minima in the vector field, and there are only saddle points.

The APF method and its variation navigation function method have been widely used in various tasks of robotic swarms, especially in the task of large-scale robotic swarm passing-through cluttered environments. In \cite{masoud2007decentralized}, the authors use the harmonic potential field function and rotational vector field to achieve the task of a robotic swarm passing through complex environments without collision among robots. In \cite{panagou2015distributed}, the authors achieve similar control performance using the navigation function method, but extend the controlled model from a single integrator model to a unicycle model. In \cite{zhao2020multi}, the authors propose a multi-robot collision avoidance algorithm based on the APF method and fuzzy strategy. In order to avoid the local minima, this algorithm optimizes the repulsive potential field function in the APF method. The fuzzy strategy is used to plan the speed of each robot to improve the passing-through efficiency.

\subsubsection{Lyapunov Guidance Vector Field Method}

The Lyapunov guidance vector field (LGVF) method is also a typical gradient vector field algorithm. This LGVF method is first proposed in \cite{frew2005cooperative} to achieve the continuous standoff tracking of fixed-wing UAVs. The LGVF method is primarily used for trajectory tracking control and target encirclement control in both single-robot and multi-robot scenarios. 

\begin{figure}[!t]
	\begin{centering}
		\includegraphics[width=\columnwidth]{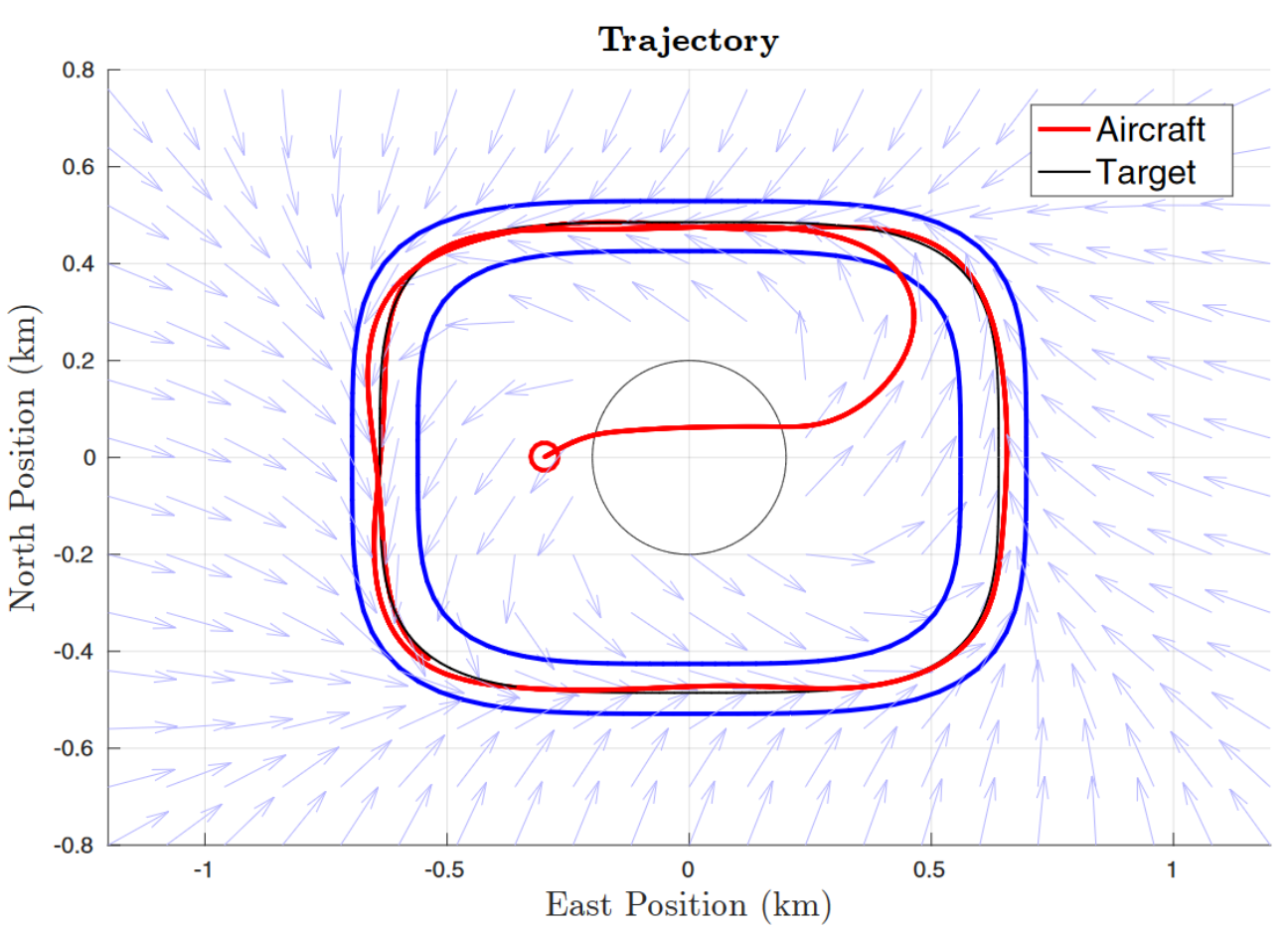}
		\par \end{centering}
	\caption{The vector field and flight trajectory generated by the Lyapunov guidance vector field method \cite{rezende2018robust}. }
	\label{VectorField}
\end{figure}

In \cite{goncalves2010vector}, the LGVF method is employed to achieve stable tracking of high-dimensional arbitrary static curves by robots. Similarly, in \cite{frew2017tracking}, the authors achieve stable tracking of dynamic star-shaped curves by robots. In \cite{rezende2021constructive}, the authors further extend the algorithm to track time-varying dynamic curves. In \cite{rezende2018robust}, the LGVF method is used to realize three-dimensional curve tracking by fixed-wing UAVs, meanwhile considering the influence of external disturbances and noises as shown in Figure \ref{VectorField}. In \cite{pothen2017curvature}, the authors address the task of sustained circular tracking of a target and employ the LGVF method to achieve stable encirclement tracking by fixed-wing UAVs. They also take into account the constraint of the turning radius. In \cite{yao2021distributed}, the authors propose improvements to the LGVF method, eliminating all singular points of the vector field while enabling cooperative curve tracking of robotic swarms. In \cite{marchidan2020collision}, the LGVF method is applied to accomplish obstacle avoidance tasks for a single fixed-wing UAV navigating through multiple static and dynamic obstacles.

\begin{figure*}[!t]
	\begin{centering}
		\includegraphics[width=0.9\textwidth]{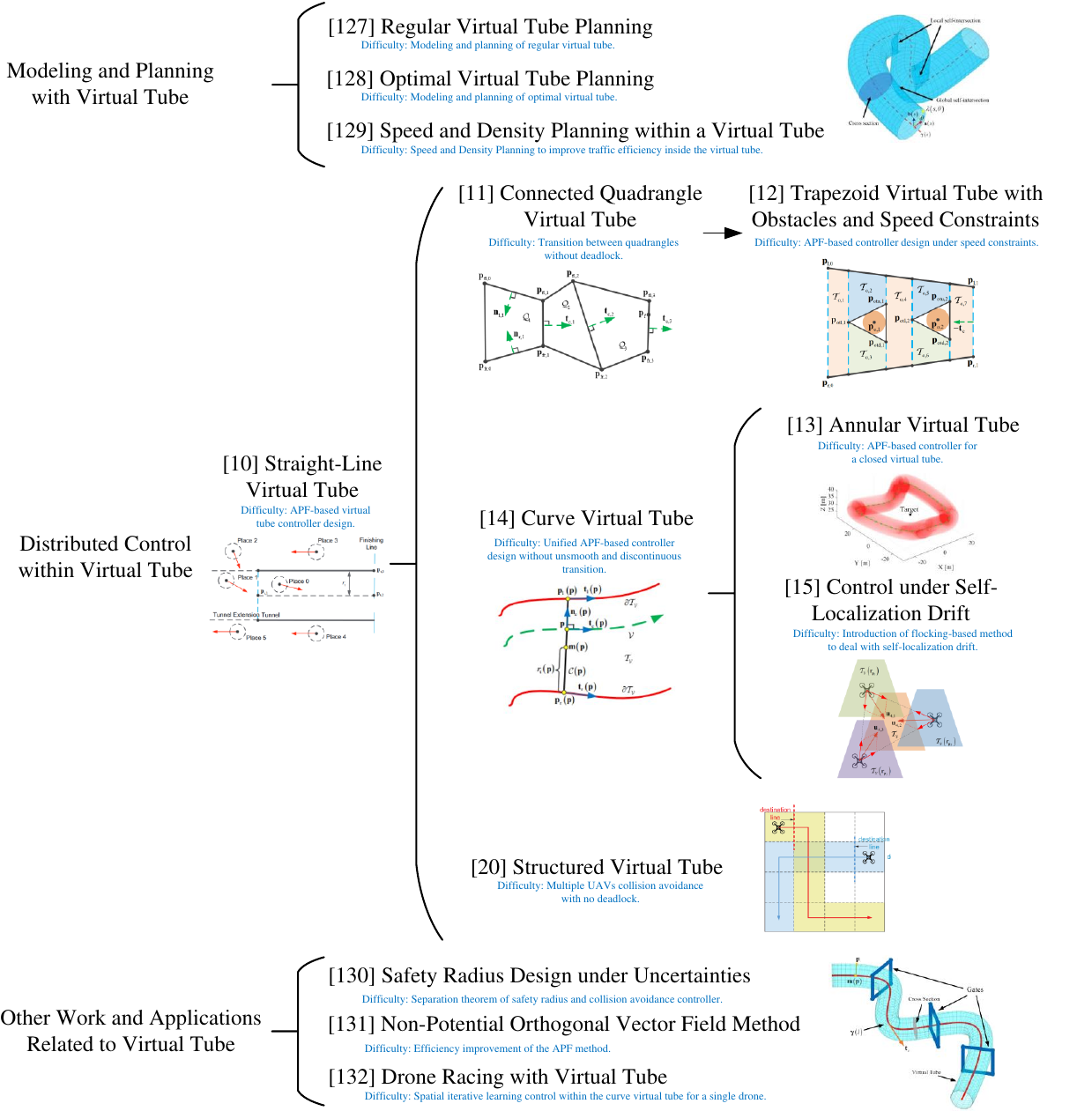}
		\par \end{centering}
	\caption{A road map of the virtual tube planning and control.}
	\label{LineMap2}
\end{figure*}

\subsubsection{Non-Gradient Vector Field Method}

There is not much research on non-gradient vector field algorithms. Due to the lack of relevant properties of potential functions, it is difficult to provide strict theoretical proofs. In \cite{panagou2014motion,panagou2016distributed}, the authors propose a non-gradient vector field algorithm for collision avoidance control among robots. The proposed algorithm can be applied to both holonomic and non-holonomic robot models. In \cite{he2022simultaneous}, the authors propose a dynamic vector field method based on the rigid body model. Specifically, the dynamics of the robot orientation are brought into the vector field, implying that the vector field is not static on the plane anymore, but a dynamic one varying with the attitude angle. In \cite{he2023novel}, the authors further focus on the motion planning for mobile robots modeled by 6-DOF rigid body systems with nonholonomic kinematics constraints. Inspired by the interaction of magnetic fields, the concept of the circular field is first introduced in \cite{singh1996real}. Robots only change their velocity direction while maintaining a constant speed. Subsequently, the papers \cite{haddadin2011dynamic, ataka2018reactive, laha2021reactive} gradually improve the circular field to adapt to different environments. Furthermore, in \cite{becker2021circular}, a circular field predictive prediction approach is designed, combining local reactive control with global motion planning, resulting in significant improvements in trajectory quality.

\subsection{Control Barrier Function}

Control barrier function (CBF) is an emerging control algorithm that has gained popularity in recent years. The primary objective of this algorithm, as described in \cite{ames2019control}, is to ensure the safety of a system. The concept of safety in control systems is first introduced in \cite{lamport1977proving}. The study of system safety is closely related to the theory of invariant sets \cite{slotine1991applied}, which has a long history of research. A comprehensive description and proof of the theory of invariant sets can be found in \cite{abraham2012manifolds}. In order to investigate the safety of a system, the CBF method divides the entire state space into three superlevel sets: the safety set, the boundary set, and the danger set.  For a dynamical system that satisfies the local Lipschitz condition, if its initial state lies within the safety set and under the control of the CBF controller, the system state remains within the safety set for an infinite time. In this case, the safety set is said to be ``forward invariant''. Additionally, the CBF method bears similarities to the Lyapunov barrier function method, which is used to address the asymptotic stability problem of control systems \cite{ames2014rapidly}. As shown in Figure \ref{Compare}, the CBF method usually has a better control performance than the flocking control and the vector field control.

To ensure system safety, the CBF method follows the principle of minimally invasive nominal control \cite{freeman2008robust}. Taking the example of multiple robots reaching their target points, the control input that guides the robots to their target positions is the nominal control input. To prevent collisions among robots, the nominal control input needs to be modified. The CBF method formulates a quadratic programming (QP) problem, where the objective function is the squared norm of the difference between the nominal control input and the modified control input. The control objective for ensuring system safety is described as linear inequality constraints containing barrier functions \cite{borrmann2015control}.

The CBF method has been widely applied in various applications in robotic swarms. In \cite{borrmann2015control}, the CBF method is employed for obstacle avoidance in a robotic swarm, where the robot model is a double integrator. Furthermore, this paper demonstrates the extension of the CBF method from centralized to distributed control, where each robot solves its own QP problem locally. In \cite{ames2014control}, the CBF method and Lyapunov barrier function method are combined for adaptive cruise control of ground autonomous vehicles, ensuring that the vehicles remain safely within their lanes. In \cite{wang2016multi}, the authors utilize the CBF method to simultaneously achieve collision avoidance and connectivity maintenance in a robotic swarm, while ensuring the non-emptiness of the feasible solution set for the QP problem. In \cite{glotfelter2017nonsmooth}, the barrier function concept is extended from smooth functions to general nonsmooth functions, and the CBF method is employed for collision avoidance among robots and with obstacles. Leveraging the differential flatness property of multicopters, the authors in \cite{wang2017safe} apply a high-order CBF method for agile control of multicopter swarms. In \cite{squires2018constructive}, the authors extend the CBF method to collision avoidance control in fixed-wing UAV swarms. Considering the nonholonomic dynamics and minimum flight speed constraints of fixed-wing UAVs, this paper designs nominal collision avoidance maneuvers. Subsequently, in \cite{squires2021safety}, this research is further extended to account for the sensing range of fixed-wing UAVs.

\section{Virtual tube Planning and Control}

The virtual tube planning and control offers a novel solution for successful guiding a multi-robot system through cluttered environments. Motivated by the AIRBUS's Skyways project \cite{AIRBUS}, the concept of the virtual tube is first proposed in \cite{quan2021practical}. To ensure safe passing-through, the robots are confined within the virtual tube and protected from potential collisions with obstacles outside the tube boundary. Therefore, the virtual tube provides a safe and hazard-free zone for robots. As shown in Figure \ref{Compare}, the virtual tube planning and control is most suitable for guiding a multi-robot system to pass through a cluttered environment. Besides, all robots can operate autonomously without wireless communication and other robots' IDs, which is rather difficult for the formation control. In Figure \ref{LineMap2}, we present a road map of the virtual tube planning and control.

\subsection{Virtual Tube with a Single Form}

\begin{figure}[!t]
	\begin{centering}
		\includegraphics[scale=0.6]{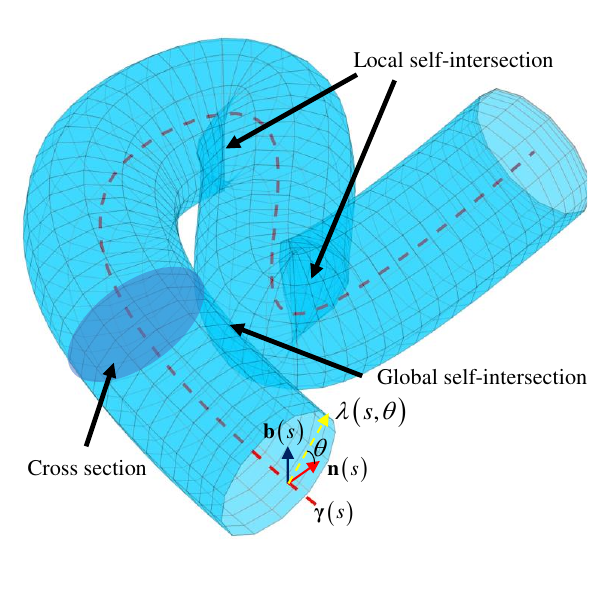}
		\par \end{centering}
	\caption{A regular virtual tube in three-dimensional space \cite{mao2022making}.}
	\label{virtual_tube_pdf}
\end{figure}

In this subsection, we introduce some kinds of virtual tubes with a single form. These virtual tubes are all \emph{regular virtual tubes}, which are satisfied with four principles.
\begin{itemize}
	\item[(1)] The virtual tube surface is regular with no block or oscillation. 
	\item[(2)] The virtual tube shrinkage rate is low.
	\item[(3)] The cross section of the virtual tube is
	simply connected.
	\item[(4)] The virtual tube has no self-intersection.
\end{itemize}
As shown in Figure \ref{virtual_tube_pdf}, the authors in \cite{mao2022making} propose a model of a class of regular virtual tubes. The authors also propose a method to obtain a regular virtual tube based on trajectory planning and regular conditions. In \cite{mao2023optimal}, the authors further propose the concept of the \emph{optimal virtual tube}, which is always regular and includes infinite optimal trajectories. Under certain conditions, any optimal trajectory in the optimal virtual tube can be expressed as a convex combination of a finite number of optimal trajectories. The authors also propose a practical planning method of the optimal virtual tube.

\begin{figure}[!t]
	\begin{centering}
		\includegraphics[scale=0.4]{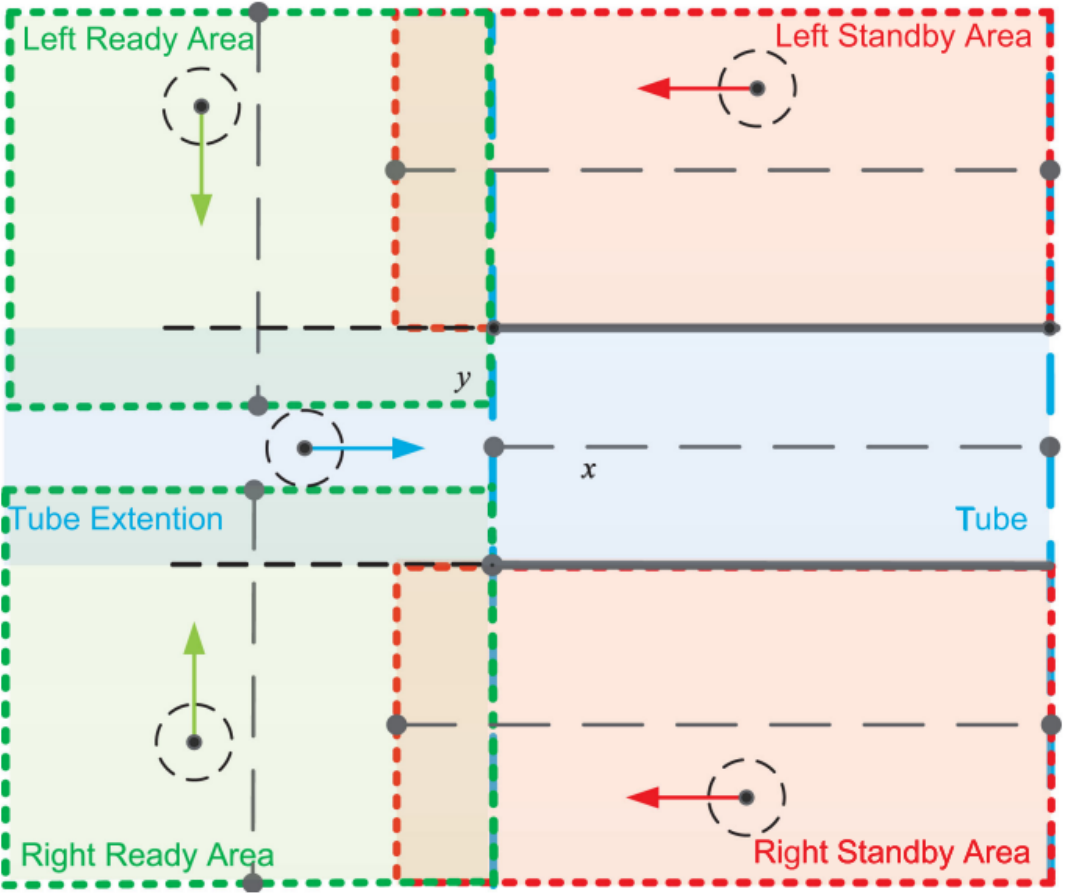}
		\par \end{centering}
	\caption{A straight-line virtual tube and some corresponding areas \cite{quan2021practical}.}
	\label{StraightLine}
\end{figure}

\begin{figure}[!t]
	\begin{centering}
		\includegraphics[width=\columnwidth]{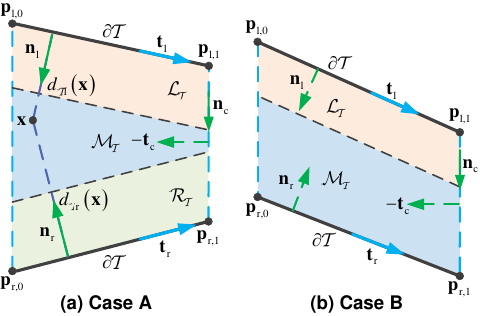}
		\par \end{centering}
	\caption{Several sub trapezoid virtual tubes for obstacle avoidance \cite{gao2022distributed2}.}
	\label{MLR}
\end{figure}

So far, many works focus on the control problem within regular virtual tubes. 
In \cite{quan2021practical}, the authors focus on distributed coordinating the motions of Vertical TakeOff and Landing (VTOL) UAVs to pass through an airway. As shown in Figure \ref{StraightLine}, the airway is modeled as a \emph{straight-line virtual tube}. By the
proposed distributed control, a VTOL UAV can keep away from another VTOL UAV or return back to the virtual tube as soon as possible, once it enters into the safety area of another or has a collision with the virtual tube while it is passing through the virtual tube. In \cite{gao2022distributed}, the authors design a \emph{trapezoid virtual tube}. Compared with the straight-line virtual tube, the width of the trapezoid virtual tube is mutable. In \cite{gao2022distributed2}, the authors generalize the application range of the trapezoid virtual tube to the condition that there exist obstacles inside and UAVs have strict speed constraints as shown in Figure \ref{MLR}. The relationship between the trapezoid virtual tube and the speed constraints is first presented. Besides, the key point of the obstacle avoidance is to divide the trapezoid virtual tube containing obstacles into several sub trapezoid virtual tubes with no obstacle inside.

\begin{figure}[!t]
	\begin{centering}
		\includegraphics[width=\columnwidth]{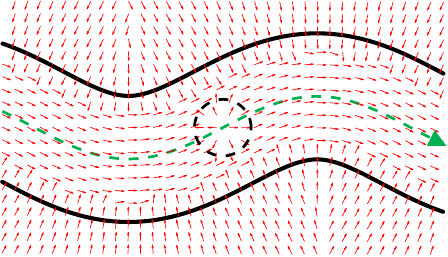}
		\par \end{centering}
	\caption{Vector field of a curve virtual tube \cite{quan2023distributed}.}
	\label{MVF}
\end{figure}

In \cite{quan2023distributed}, the authors design a \emph{curve virtual tube} for the robotic swarm as shown in Figure \ref{MVF}. The considered curve virtual tube is always \emph{regular}. To guide a robotic swarm within the curve virtual tube, a distributed vector field controller is proposed with three elaborate control terms. For convenience in practical use, a modified controller with an approximate control performance is also put forward. In \cite{gao2022robust}, the authors generalize the controller design inside the curve virtual tube to the condition that all robots have self-localization drifts and precise relative navigation, where the flocking algorithm is introduced to reduce the negative impact of the self-localization drift. Similar to the ``many wrongs principle'', the cohesion behavior and the velocity alignment behavior are able to reduce the influence of the position measurement drift and the velocity measurement error, respectively. In \cite{song2023speed}, based on a planned curve virtual tube, the average forward speed and density along the virtual tube are further planned to ensure safety and improve efficiency for a large number of speed-constrained robots. In \cite{gao2022multi}, the authors design an \emph{annular virtual tube}, which can be seen as a closed form of the curve virtual tube. Instead of the traditional methods of all UAVs converging to a closed curve, the authors let all UAVs converge to an annular virtual tube and achieve the target encirclement.

\subsection{Virtual Tube with a Combined Form}
\begin{figure}[!t]
	\begin{centering}
		\includegraphics[width=0.9\columnwidth]{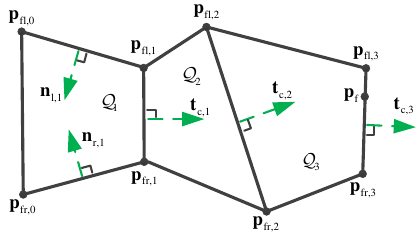}
		\par \end{centering}
	\caption{A connected quadrangle virtual tube \cite{gao2022distributed}. }
	\label{TubePF}
\end{figure}

\begin{figure}[!t]
	\begin{centering}
		\includegraphics[width=0.9\columnwidth]{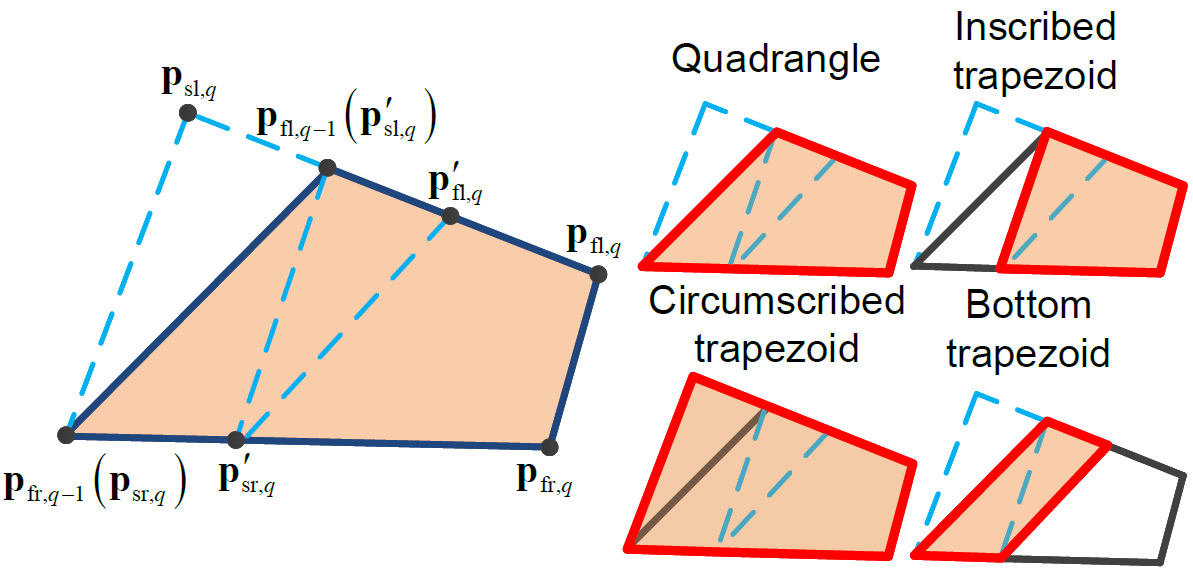}
		\par \end{centering}
	\caption{Inscribed trapezoid, circumscribed trapezoid and bottom trapezoid of
		a quadrangle \cite{gao2022distributed}.}
	\label{inscribe}
\end{figure}

In this subsection, we introduce some kinds of virtual tubes with a combined form. In \cite{gao2022distributed}, the authors design a \emph{connected quadrangle virtual tube} as shown in Figure \ref{TubePF}. The basis of the connected quadrangle virtual tube is the trapezoid virtual tube. For the connected quadrangle virtual tube, several corresponding trapezoid virtual tubes and a modified switching logic are proposed to avoid the deadlock and prevent agents from moving outside the virtual tube as shown in Figure \ref{inscribe}. In \cite{fu2022practical}, the authors design a \emph{structured virtual tube} to solve the free flight control problem, which includes convergence to destination lines/planes and inter-agent collision avoidance. Based on the structured virtual tube, the authors in \cite{safadi2023macroscopic} investigate the collective and aggregate aircraft traffic flow diagrams for low-altitude air city transport systems. In \cite{quan2021sky}, a ``sky highway'' is designed based on virtual tubes and rotary islands. In the sky highway, each UAV will have its route, and an airway like a highway road can allow many UAVs to perform free flight simultaneously.

\subsection{Other Work and Applications Related to Virtual Tube}
When robots are moving inside a virtual tube, collision avoidance among robots is very important. In \cite{quan2022far}, the authors propose a separation principle of the safety radius design and controller design subject to communication uncertainties. With the separation principle, the safety radius in the design phase (without uncertainties) and flight phase (subject to uncertainties) are studied. This work gives a guideline of how far two UAVs should be when they are passing through a cluttered environment. 

High traffic efficiency is usually a control objective inside the virtual tube. In \cite{gao2023non}, the authors propose a non-potential orthogonal vector field method, which is the modification of the classic attractive/repulsive potential fields approach so as to improve its efficiency while retaining the Lyapunov stability analysis from traditional potential fields. The improvement strategy aims at making the overall repulsive vector field orthogonal to the attractive vector field in some conditions. This modification allows the robot to move faster toward the goal since the impact of the repulsive potential field is reduced in the closed-loop system. 

\begin{figure}[!t]
	\begin{centering}
		\includegraphics[width=0.8\columnwidth]{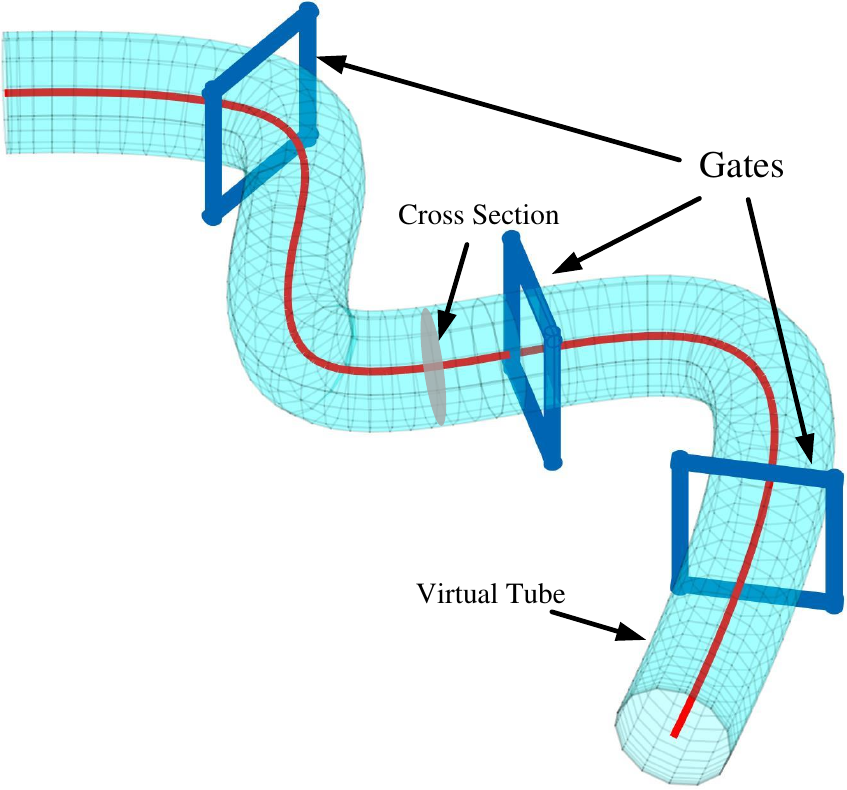}
		\par \end{centering}
	\caption{Virtual tube suitable for the racing track \cite{lv2023autonomous}. }
	\label{tuberace1}
\end{figure}

Besides the robotic swarm, the virtual tube can also be applied to some applications of a single robot. In \cite{lv2023autonomous}, the concept of the virtual tube is introduced to autonomous drone race as shown in Figure \ref{tuberace1}. The authors develop a highly efficient learning method by imitating the training experience of racing drivers. Unlike traditional iterative learning control methods for accurate tracking, the proposed approach iteratively learns a trajectory online to finish the race as quickly as possible.

\subsection{Further Work on Virtual Tubes}

\begin{figure}[!t]
	\begin{centering}
		\includegraphics[width=\columnwidth]{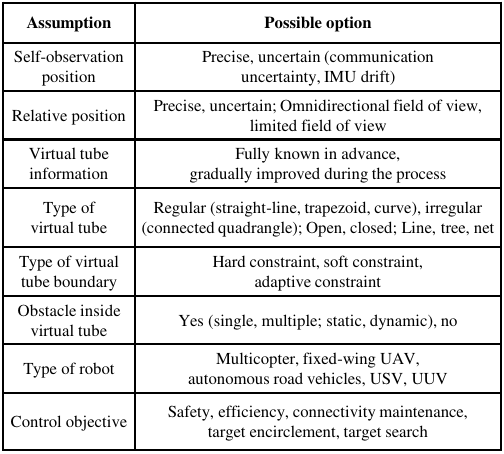}
		\par \end{centering}
	\caption{Further work on virtual tube planning and control.}
	\label{Future}
\end{figure}

As a novel solution for guiding a multi-robot system through cluttered environments, the virtual tube has many interesting topics to be investigated. For example, when robots have fan-shaped detection areas like Figure \ref{DetectionArea}, the collision avoidance algorithm inside the virtual tube should be modified. Besides, the sky highway can be seen as virtual tubes with net forms. How to solve the planning and control problem within the sky highway needs to be investigated. In Figure \ref{Future}, the assumptions are classified into eight categories, and each type of assumption has several possible options. For example, the work \cite{quan2023distributed} corresponds to that the self-observation position and relative position are precise, the virtual tube information is fully known in advance, the type of virtual tube is curve virtual tube, the type of virtual tube boundary is hard constraint, no obstacle is inside the virtual tube, the type of robot is quadcopter, and the control objective is safety. By choosing some of these assumptions and options, we can get several meaningful topics for our future work.

\section{Conclusion}

In this survey, we have provided a comprehensive review of various methods and algorithms related to passing-through control of multi-robot systems in cluttered environments. This continues to be an active area of research, and we highlight a number of channels where current approaches may be improved. We first describe some models of robots and commonly considered control objectives. Then we introduce four types of algorithms that can be employed for passing-through control. We provide some subjective and general evaluations. At last, we also point out some meaningful future work related to virtual tubes.

\bibliographystyle{IEEEtran}
\bibliography{survey}

\begin{thebibliography}{100}
\providecommand{\url}[1]{#1}
\csname url@samestyle\endcsname
\providecommand{\newblock}{\relax}
\providecommand{\bibinfo}[2]{#2}
\providecommand{\BIBentrySTDinterwordspacing}{\spaceskip=0pt\relax}
\providecommand{\BIBentryALTinterwordstretchfactor}{4}
\providecommand{\BIBentryALTinterwordspacing}{\spaceskip=\fontdimen2\font plus
\BIBentryALTinterwordstretchfactor\fontdimen3\font minus
  \fontdimen4\font\relax}
\providecommand{\BIBforeignlanguage}[2]{{%
\expandafter\ifx\csname l@#1\endcsname\relax
\typeout{** WARNING: IEEEtran.bst: No hyphenation pattern has been}%
\typeout{** loaded for the language `#1'. Using the pattern for}%
\typeout{** the default language instead.}%
\else
\language=\csname l@#1\endcsname
\fi
#2}}
\providecommand{\BIBdecl}{\relax}
\BIBdecl

\bibitem{oh2015survey}
K.-K. Oh, M.-C. Park, and H.-S. Ahn, ``A survey of multi-agent formation
  control,'' \emph{Automatica}, vol.~53, pp. 424--440, 2015.

\bibitem{jiang2022bibliometric}
Y.~Jiang, Y.~Gao, W.~Song, Y.~Li, and Q.~Quan, ``Bibliometric analysis of uav
  swarms,'' \emph{Journal of Systems Engineering and Electronics}, vol.~33,
  no.~2, pp. 406--425, 2022.

\bibitem{do2021formation}
H.~T. Do, H.~T. Hua, M.~T. Nguyen, C.~V. Nguyen, H.~T. Nguyen, H.~T. Nguyen,
  and N.~T. Nguyen, ``Formation control algorithms for multiple-uavs: a
  comprehensive survey,'' \emph{EAI Endorsed Transactions on Industrial
  Networks and Intelligent Systems}, vol.~8, no.~27, pp. e3--e3, 2021.

\bibitem{chen2020dynamic}
X.~Chen, M.~Zhao, and L.~Yin, ``Dynamic path planning of the uav avoiding
  static and moving obstacles,'' \emph{Journal of Intelligent \& Robotic
  Systems}, vol.~99, pp. 909--931, 2020.

\bibitem{askari2016new}
A.~Askari, M.~Mortazavi, H.~Talebi, and A.~Motamedi, ``A new approach in uav
  path planning using bezier--dubins continuous curvature path,''
  \emph{Proceedings of the Institution of Mechanical Engineers, Part G: Journal
  of Aerospace Engineering}, vol. 230, no.~6, pp. 1103--1113, 2016.

\bibitem{zhou2020robust}
B.~Zhou, F.~Gao, J.~Pan, and S.~Shen, ``Robust real-time uav replanning using
  guided gradient-based optimization and topological paths,'' in \emph{2020
  IEEE International Conference on Robotics and Automation (ICRA)}.\hskip 1em
  plus 0.5em minus 0.4em\relax IEEE, 2020, pp. 1208--1214.

\bibitem{khatib1986real}
O.~Khatib, ``Real-time obstacle avoidance for manipulators and mobile robots,''
  \emph{The International Journal of Robotics Research}, vol.~5, no.~1, pp.
  90--98, 1986.

\bibitem{kim1992real}
J.-O. Kim and P.~K. Khosla, ``Real-time obstacle avoidance using harmonic
  potential functions,'' \emph{IEEE Transactions on Robotics and Automation},
  vol.~8, no. 3 Ju, 1992.

\bibitem{rimon1990exact}
E.~Rimon, \emph{Exact robot navigation using artificial potential
  functions}.\hskip 1em plus 0.5em minus 0.4em\relax Yale University, 1990.

\bibitem{quan2021practical}
Q.~Quan, R.~Fu, M.~Li, D.~Wei, Y.~Gao, and K.-Y. Cai, ``Practical distributed
  control for {VTOL UAVs} to pass a virtual tube,'' \emph{IEEE Transactions on
  Intelligent Vehicles}, vol.~7, no.~2, pp. 342--353, 2021.

\bibitem{gao2022distributed}
Y.~Gao, C.~Bai, and Q.~Quan, ``Distributed control for a multi-agent system to
  pass through a connected quadrangle virtual tube,'' \emph{IEEE Transactions
  on Control of Network Systems}, vol.~10, no.~2, pp. 693--705, 2023.

\bibitem{gao2022distributed2}
------, ``Distributed control within a trapezoid virtual tube containing
  obstacles for uav swarm subject to speed constraints,'' \emph{arXiv preprint
  arXiv:2212.12640}, 2022.

\bibitem{gao2022multi}
Y.~Gao, C.~Bai, L.~Zhang, and Q.~Quan, ``Multi-uav cooperative target
  encirclement within an annular virtual tube,'' \emph{Aerospace Science and
  Technology}, vol. 128, p. 107800, 2022.

\bibitem{quan2023distributed}
Q.~Quan, Y.~Gao, and C.~Bai, ``Distributed control for a robotic swarm to pass
  through a curve virtual tube,'' \emph{Robotics and Autonomous Systems}, p.
  104368, 2023.

\bibitem{gao2022robust}
Y.~Gao, C.~Bai, and Q.~Quan, ``Robust distributed control within a curve
  virtual tube for a robotic swarm under self-localization drift and precise
  relative navigation,'' \emph{International Journal of Robust and Nonlinear
  Control}, 2023.

\bibitem{rasekhipour2016potential}
Y.~Rasekhipour, A.~Khajepour, S.-K. Chen, and B.~Litkouhi, ``A potential
  field-based model predictive path-planning controller for autonomous road
  vehicles,'' \emph{IEEE Transactions on Intelligent Transportation Systems},
  vol.~18, no.~5, pp. 1255--1267, 2016.

\bibitem{luo2018porca}
Y.~Luo, P.~Cai, A.~Bera, D.~Hsu, W.~S. Lee, and D.~Manocha, ``Porca: Modeling
  and planning for autonomous driving among many pedestrians,'' \emph{IEEE
  Robotics and Automation Letters}, vol.~3, no.~4, pp. 3418--3425, 2018.

\bibitem{liu2017planning}
S.~Liu, M.~Watterson, K.~Mohta, K.~Sun, S.~Bhattacharya, C.~J. Taylor, and
  V.~Kumar, ``Planning dynamically feasible trajectories for quadrotors using
  safe flight corridors in 3-d complex environments,'' \emph{IEEE Robotics and
  Automation Letters}, vol.~2, no.~3, pp. 1688--1695, 2017.

\bibitem{quan2021sky}
Q.~Quan, M.~Li, and R.~Fu, ``Sky highway design for dense traffic,''
  \emph{IFAC-PapersOnLine}, vol.~54, no.~2, pp. 140--145, 2021.

\bibitem{fu2022practical}
R.~Fu, Q.~Quan, M.~Li, and K.-Y. Cai, ``Practical distributed control for
  cooperative multicopters in structured free flight concepts,'' \emph{IEEE
  Transactions on Intelligent Transportation Systems}, vol.~24, no.~4, pp.
  4203--4216, 2022.

\bibitem{safadi2023macroscopic}
Y.~Safadi, R.~Fu, Q.~Quan, and J.~Haddad, ``Macroscopic fundamental diagrams
  for low-altitude air city transport,'' \emph{Transportation Research Part C:
  Emerging Technologies}, vol. 152, p. 104141, 2023.

\bibitem{dai2021rflysim}
X.~Dai, C.~Ke, Q.~Quan, and K.-Y. Cai, ``Rflysim: Automatic test platform for
  {UAV} autopilot systems with fpga-based hardware-in-the-loop simulations,''
  \emph{Aerospace Science and Technology}, vol. 114, p. 106727, 2021.

\bibitem{wang2021rflysim}
S.~Wang, X.~Dai, C.~Ke, and Q.~Quan, ``Rflysim: A rapid multicopter development
  platform for education and research based on pixhawk and {MATLAB},'' in
  \emph{2021 International Conference on Unmanned Aircraft Systems
  (ICUAS)}.\hskip 1em plus 0.5em minus 0.4em\relax IEEE, 2021, pp. 1587--1594.

\bibitem{shah2018airsim}
S.~Shah, D.~Dey, C.~Lovett, and A.~Kapoor, ``Airsim: High-fidelity visual and
  physical simulation for autonomous vehicles,'' in \emph{Field and Service
  Robotics: Results of the 11th International Conference}.\hskip 1em plus 0.5em
  minus 0.4em\relax Springer, 2018, pp. 621--635.

\bibitem{liu2020multi}
D.~Liu, C.~Zong, D.~Wang, W.~Zhao, Y.~Wang, and W.~Lu, ``Multi-robot formation
  control based on high-order bilateral consensus,'' \emph{Measurement and
  Control}, vol.~53, no. 5-6, pp. 983--993, 2020.

\bibitem{hoy2015algorithms}
M.~Hoy, A.~S. Matveev, and A.~V. Savkin, ``Algorithms for collision-free
  navigation of mobile robots in complex cluttered environments: a survey,''
  \emph{Robotica}, vol.~33, no.~3, pp. 463--497, 2015.

\bibitem{hu2014adaptive}
J.~Hu and W.~X. Zheng, ``Adaptive tracking control of leader--follower systems
  with unknown dynamics and partial measurements,'' \emph{Automatica}, vol.~50,
  no.~5, pp. 1416--1423, 2014.

\bibitem{zhao2018affine}
S.~Zhao, ``Affine formation maneuver control of multiagent systems,''
  \emph{IEEE Transactions on Automatic Control}, vol.~63, no.~12, pp.
  4140--4155, 2018.

\bibitem{chen2020distributed}
L.~Chen, J.~Mei, C.~Li, and G.~Ma, ``Distributed leader--follower affine
  formation maneuver control for high-order multiagent systems,'' \emph{IEEE
  Transactions on Automatic Control}, vol.~65, no.~11, pp. 4941--4948, 2020.

\bibitem{xu2019affine}
Y.~Xu, D.~Luo, D.~Li, Y.~You, and H.~Duan, ``Affine formation control for
  heterogeneous multi-agent systems with directed interaction networks,''
  \emph{Neurocomputing}, vol. 330, pp. 104--115, 2019.

\bibitem{anderson2008rigid}
B.~D. Anderson, C.~Yu, B.~Fidan, and J.~M. Hendrickx, ``Rigid graph control
  architectures for autonomous formations,'' \emph{IEEE Control Systems
  Magazine}, vol.~28, no.~6, pp. 48--63, 2008.

\bibitem{krick2009stabilisation}
L.~Krick, M.~E. Broucke, and B.~A. Francis, ``Stabilisation of infinitesimally
  rigid formations of multi-robot networks,'' \emph{International Journal of
  Control}, vol.~82, no.~3, pp. 423--439, 2009.

\bibitem{rozenheck2015proportional}
O.~Rozenheck, S.~Zhao, and D.~Zelazo, ``A proportional-integral controller for
  distance-based formation tracking,'' in \emph{2015 European Control
  Conference (ECC)}.\hskip 1em plus 0.5em minus 0.4em\relax Austria: IEEE,
  2015, pp. 1693--1698.

\bibitem{bae2020distributed}
Y.~Bae, Y.~Lim, and H.~Ahn, ``Distributed robust adaptive gradient controller
  in distance-based formation control with exogenous disturbance,'' \emph{IEEE
  Transactions on Automatic Control}, vol.~66, no.~6, pp. 2868--2874, 2020.

\bibitem{zhao2019bearing}
S.~Zhao and D.~Zelazo, ``Bearing rigidity theory and its applications for
  control and estimation of network systems: Life beyond distance rigidity,''
  \emph{IEEE Control Systems Magazine}, vol.~39, no.~2, pp. 66--83, 2019.

\bibitem{zhao2015bearing}
------, ``Bearing rigidity and almost global bearing-only formation
  stabilization,'' \emph{IEEE Transactions on Automatic Control}, vol.~61,
  no.~5, pp. 1255--1268, 2015.

\bibitem{li2020adaptive}
X.~Li, C.~Wen, and C.~Chen, ``Adaptive formation control of networked robotic
  systems with bearing-only measurements,'' \emph{IEEE Transactions on
  Cybernetics}, vol.~51, no.~1, pp. 199--209, 2020.

\bibitem{dijkstra2022note}
E.~W. Dijkstra, ``A note on two problems in connexion with graphs,'' in
  \emph{Edsger Wybe Dijkstra: His Life, Work, and Legacy}, 2022, pp. 287--290.

\bibitem{hart1968formal}
P.~E. Hart, N.~J. Nilsson, and B.~Raphael, ``A formal basis for the heuristic
  determination of minimum cost paths,'' \emph{IEEE transactions on Systems
  Science and Cybernetics}, vol.~4, no.~2, pp. 100--107, 1968.

\bibitem{likhachev2003ara}
M.~Likhachev, G.~J. Gordon, and S.~Thrun, ``Ara*: Anytime a* with provable
  bounds on sub-optimality,'' \emph{Advances in neural information processing
  systems}, vol.~16, 2003.

\bibitem{dolgov2008practical}
D.~Dolgov, S.~Thrun, M.~Montemerlo, and J.~Diebel, ``Practical search
  techniques in path planning for autonomous driving,'' \emph{Ann Arbor}, vol.
  1001, no. 48105, pp. 18--80, 2008.

\bibitem{harabor2011online}
D.~Harabor and A.~Grastien, ``Online graph pruning for pathfinding on grid
  maps,'' in \emph{Proceedings of the AAAI Conference on Artificial
  Intelligence}, vol.~25, no.~1, 2011, pp. 1114--1119.

\bibitem{geraerts2004comparative}
R.~Geraerts and M.~H. Overmars, ``A comparative study of probabilistic roadmap
  planners,'' \emph{Algorithmic foundations of robotics V}, pp. 43--57, 2004.

\bibitem{lavalle2001randomized}
S.~M. LaValle and J.~J. Kuffner~Jr, ``Randomized kinodynamic planning,''
  \emph{The international journal of robotics research}, vol.~20, no.~5, pp.
  378--400, 2001.

\bibitem{karaman2010incremental}
S.~Karaman and E.~Frazzoli, ``Incremental sampling-based algorithms for optimal
  motion planning,'' \emph{Robotics Science and Systems VI}, vol. 104, no.~2,
  2010.

\bibitem{patle2019review}
B.~Patle, A.~Pandey, D.~Parhi, A.~Jagadeesh \emph{et~al.}, ``A review: On path
  planning strategies for navigation of mobile robot,'' \emph{Defence
  Technology}, vol.~15, no.~4, pp. 582--606, 2019.

\bibitem{aggarwal2020path}
S.~Aggarwal and N.~Kumar, ``Path planning techniques for unmanned aerial
  vehicles: A review, solutions, and challenges,'' \emph{Computer
  Communications}, vol. 149, pp. 270--299, 2020.

\bibitem{cheng2021path}
C.~Cheng, Q.~Sha, B.~He, and G.~Li, ``Path planning and obstacle avoidance for
  auv: A review,'' \emph{Ocean Engineering}, vol. 235, p. 109355, 2021.

\bibitem{wang2022geometrically}
Z.~Wang, X.~Zhou, C.~Xu, and F.~Gao, ``Geometrically constrained trajectory
  optimization for multicopters,'' \emph{IEEE Transactions on Robotics},
  vol.~38, no.~5, pp. 3259--3278, 2022.

\bibitem{mellinger2012mixed}
D.~Mellinger, A.~Kushleyev, and V.~Kumar, ``Mixed-integer quadratic program
  trajectory generation for heterogeneous quadrotor teams,'' in \emph{2012 IEEE
  International Conference on Robotics and Automation}.\hskip 1em plus 0.5em
  minus 0.4em\relax IEEE, 2012, pp. 477--483.

\bibitem{culligan2006online}
K.~F. Culligan, ``Online trajectory planning for uavs using mixed integer
  linear programming,'' Ph.D. dissertation, Massachusetts Institute of
  Technology, 2006.

\bibitem{van1998real}
M.~J. Van~Nieuwstadt and R.~M. Murray, ``Real-time trajectory generation for
  differentially flat systems,'' \emph{International Journal of Robust and
  Nonlinear Control: IFAC-Affiliated Journal}, vol.~8, no.~11, pp. 995--1020,
  1998.

\bibitem{mellinger2011minimum}
D.~Mellinger and V.~Kumar, ``Minimum snap trajectory generation and control for
  quadrotors,'' in \emph{2011 IEEE international Conference on Robotics and
  Automation}.\hskip 1em plus 0.5em minus 0.4em\relax IEEE, 2011, pp.
  2520--2525.

\bibitem{richter2016polynomial}
C.~Richter, A.~Bry, and N.~Roy, ``Polynomial trajectory planning for aggressive
  quadrotor flight in dense indoor environments,'' in \emph{Robotics Research:
  The 16th International Symposium ISRR}.\hskip 1em plus 0.5em minus
  0.4em\relax Springer, 2016, pp. 649--666.

\bibitem{chen2016online}
J.~Chen, T.~Liu, and S.~Shen, ``Online generation of collision-free
  trajectories for quadrotor flight in unknown cluttered environments,'' in
  \emph{2016 IEEE International Conference on Robotics and Automation
  (ICRA)}.\hskip 1em plus 0.5em minus 0.4em\relax IEEE, 2016, pp. 1476--1483.

\bibitem{mac2017hierarchical}
T.~T. Mac, C.~Copot, D.~T. Tran, and R.~De~Keyser, ``A hierarchical global path
  planning approach for mobile robots based on multi-objective particle swarm
  optimization,'' \emph{Applied Soft Computing}, vol.~59, pp. 68--76, 2017.

\bibitem{bai2019distributed}
X.~Bai, W.~Yan, M.~Cao, and D.~Xue, ``Distributed multi-vehicle task assignment
  in a time-invariant drift field with obstacles,'' \emph{IET Control Theory \&
  Applications}, vol.~13, no.~17, pp. 2886--2893, 2019.

\bibitem{erokhin2019optimal}
A.~Erokhin, V.~Erokhin, S.~Sotnikov, and A.~Gogolevsky, ``Optimal multi-robot
  path finding algorithm based on {A*},'' in \emph{Intelligent Systems in
  Cybernetics and Automation Control Theory 2}.\hskip 1em plus 0.5em minus
  0.4em\relax Springer, 2019, pp. 172--182.

\bibitem{sun2019novel}
G.~Sun, R.~Zhou, B.~Di, Z.~Dong, and Y.~Wang, ``A novel cooperative path
  planning for multi-robot persistent coverage with obstacles and coverage
  period constraints,'' \emph{Sensors}, vol.~19, no.~9, p. 1994, 2019.

\bibitem{preiss2017downwash}
J.~A. Preiss, W.~H{\"o}nig, N.~Ayanian, and G.~S. Sukhatme, ``Downwash-aware
  trajectory planning for large quadrotor teams,'' in \emph{2017 IEEE/RSJ
  International Conference on Intelligent Robots and Systems (IROS)}.\hskip 1em
  plus 0.5em minus 0.4em\relax IEEE, 2017, pp. 250--257.

\bibitem{madridano20203d}
{\'A}.~Madridano, A.~Al-Kaff, D.~Mart{\'\i}n, and A.~de~la Escalera, ``3d
  trajectory planning method for {UAVs} swarm in building emergencies,''
  \emph{Sensors}, vol.~20, no.~3, p. 642, 2020.

\bibitem{wu2020probabilistically}
P.~Wu, L.~Li, J.~Xie, and J.~Chen, ``Probabilistically guaranteed path planning
  for safe urban air mobility using chance constrained {RRT},'' in \emph{AIAA
  Aviation 2020 Forum}, 2020, p. 2914.

\bibitem{berning2020rapid}
A.~W. Berning, A.~Girard, I.~Kolmanovsky, and S.~N. D'Souza, ``Rapid
  uncertainty propagation and chance-constrained path planning for small
  unmanned aerial vehicles,'' \emph{Advanced Control for Applications:
  Engineering and Industrial Systems}, vol.~2, no.~1, p. e23, 2020.

\bibitem{song2016rolling}
B.~D. Song, J.~Kim, and J.~R. Morrison, ``Rolling horizon path planning of an
  autonomous system of uavs for persistent cooperative service: Milp
  formulation and efficient heuristics,'' \emph{Journal of Intelligent \&
  Robotic Systems}, vol.~84, pp. 241--258, 2016.

\bibitem{lal2017optimal}
R.~Lal, A.~Sharda, and P.~Prabhakar, ``Optimal multi-robot path planning for
  pesticide spraying in agricultural fields,'' in \emph{2017 IEEE 56th Annual
  Conference on Decision and Control (CDC)}.\hskip 1em plus 0.5em minus
  0.4em\relax IEEE, 2017, pp. 5815--5820.

\bibitem{kushleyev2013towards}
A.~Kushleyev, D.~Mellinger, C.~Powers, and V.~Kumar, ``Towards a swarm of agile
  micro quadrotors,'' \emph{Autonomous Robots}, vol.~35, no.~4, pp. 287--300,
  2013.

\bibitem{soria2021predictive}
E.~Soria, F.~Schiano, and D.~Floreano, ``Predictive control of aerial swarms in
  cluttered environments,'' \emph{Nature Machine Intelligence}, vol.~3, no.~6,
  pp. 545--554, 2021.

\bibitem{zhang2019trajectory}
C.~Zhang, D.~Chu, S.~Liu, Z.~Deng, C.~Wu, and X.~Su, ``Trajectory planning and
  tracking for autonomous vehicle based on state lattice and model predictive
  control,'' \emph{IEEE Intelligent Transportation systems magazine}, vol.~11,
  no.~2, pp. 29--40, 2019.

\bibitem{park2020online}
J.~Park and H.~J. Kim, ``Online trajectory planning for multiple quadrotors in
  dynamic environments using relative safe flight corridor,'' \emph{IEEE
  Robotics and Automation Letters}, vol.~6, no.~2, pp. 659--666, 2020.

\bibitem{park2019fast}
------, ``Fast trajectory planning for multiple quadrotors using relative safe
  flight corridor,'' in \emph{2019 IEEE/RSJ International Conference on
  Intelligent Robots and Systems (IROS)}.\hskip 1em plus 0.5em minus
  0.4em\relax IEEE, 2019, pp. 596--603.

\bibitem{park2022online}
J.~Park, D.~Kim, G.~C. Kim, D.~Oh, and H.~J. Kim, ``Online distributed
  trajectory planning for quadrotor swarm with feasibility guarantee using
  linear safe corridor,'' \emph{IEEE Robotics and Automation Letters}, vol.~7,
  no.~2, pp. 4869--4876, 2022.

\bibitem{luis2020online}
C.~E. Luis, M.~Vukosavljev, and A.~P. Schoellig, ``Online trajectory generation
  with distributed model predictive control for multi-robot motion planning,''
  \emph{IEEE Robotics and Automation Letters}, vol.~5, no.~2, pp. 604--611,
  2020.

\bibitem{zhou2021ego}
X.~Zhou, J.~Zhu, H.~Zhou, C.~Xu, and F.~Gao, ``Ego-swarm: A fully autonomous
  and decentralized quadrotor swarm system in cluttered environments,'' in
  \emph{2021 IEEE international conference on robotics and automation
  (ICRA)}.\hskip 1em plus 0.5em minus 0.4em\relax IEEE, 2021, pp. 4101--4107.

\bibitem{zhou2021decentralized}
X.~Zhou, Z.~Wang, X.~Wen, J.~Zhu, C.~Xu, and F.~Gao, ``Decentralized
  spatial-temporal trajectory planning for multicopter swarms,'' \emph{arXiv
  preprint arXiv:2106.12481}, 2021.

\bibitem{reynolds1987flocks}
C.~W. Reynolds, ``Flocks, herds and schools: A distributed behavioral model,''
  in \emph{Proceedings of the 14th annual conference on Computer graphics and
  interactive techniques}, 1987, pp. 25--34.

\bibitem{liu2023task}
B.~Liu, S.~Wang, Q.~Li, X.~Zhao, Y.~Pan, and C.~Wang, ``Task assignment of uav
  swarms based on deep reinforcement learning,'' \emph{Drones}, vol.~7, no.~5,
  p. 297, 2023.

\bibitem{vicsek1995novel}
T.~Vicsek, A.~Czir{\'o}k, E.~Ben-Jacob, I.~Cohen, and O.~Shochet, ``Novel type
  of phase transition in a system of self-driven particles,'' \emph{Physical
  review letters}, vol.~75, no.~6, p. 1226, 1995.

\bibitem{couzin2002collective}
I.~D. Couzin, J.~Krause, R.~James, G.~D. Ruxton, and N.~R. Franks, ``Collective
  memory and spatial sorting in animal groups,'' \emph{Journal of theoretical
  biology}, vol. 218, no.~1, pp. 1--11, 2002.

\bibitem{cucker2007emergent}
F.~Cucker and S.~Smale, ``Emergent behavior in flocks,'' \emph{IEEE
  Transactions on Automatic Control}, vol.~52, no.~5, pp. 852--862, 2007.

\bibitem{la2011flocking}
H.~M. La and W.~Sheng, ``Flocking control algorithms for multiple agents in
  cluttered and noisy environments,'' \emph{Bio-Inspired Self-Organizing
  Robotic Systems}, pp. 53--79, 2011.

\bibitem{ban2021self}
Z.~Ban, J.~Hu, B.~Lennox, and F.~Arvin, ``Self-organised collision-free
  flocking mechanism in heterogeneous robot swarms,'' \emph{Mobile Networks and
  Applications}, pp. 1--11, 2021.

\bibitem{beaver2020optimal}
L.~E. Beaver, C.~Kroninger, and A.~A. Malikopoulos, ``An optimal control
  approach to flocking,'' in \emph{2020 American Control Conference
  (ACC)}.\hskip 1em plus 0.5em minus 0.4em\relax IEEE, 2020, pp. 683--688.

\bibitem{beaver2020beyond}
L.~E. Beaver and A.~A. Malikopoulos, ``Beyond reynolds: a constraint-driven
  approach to cluster flocking,'' in \emph{2020 59th IEEE Conference on
  Decision and Control (CDC)}.\hskip 1em plus 0.5em minus 0.4em\relax IEEE,
  2020, pp. 208--213.

\bibitem{liu2021hierarchical}
X.~Liu, X.~Xiang, Y.~Chang, C.~Yan, H.~Zhou, and D.~Tang, ``Hierarchical
  weighting vicsek model for flocking navigation of drones,'' \emph{Drones},
  vol.~5, no.~3, p.~74, 2021.

\bibitem{arfken1999mathematical}
G.~B. Arfken and H.~J. Weber, ``Mathematical methods for physicists,'' 1999.

\bibitem{panagou2014motion}
D.~Panagou, ``Motion planning and collision avoidance using navigation vector
  fields,'' in \emph{2014 IEEE International Conference on Robotics and
  Automation (ICRA)}.\hskip 1em plus 0.5em minus 0.4em\relax IEEE, 2014, pp.
  2513--2518.

\bibitem{hernandez2011convergence}
E.~G. Hern{\'a}ndez-Mart{\'\i}nez and E.~Aranda-Bricaire, \emph{Convergence and
  collision avoidance in formation control: A survey of the artificial
  potential functions approach}.\hskip 1em plus 0.5em minus 0.4em\relax INTECH
  Open Access Publisher Rijeka, Croatia, 2011.

\bibitem{rostami2019obstacle}
S.~M.~H. Rostami, A.~K. Sangaiah, J.~Wang, and X.~Liu, ``Obstacle avoidance of
  mobile robots using modified artificial potential field algorithm,''
  \emph{EURASIP Journal on Wireless Communications and Networking}, vol. 2019,
  no.~1, pp. 1--19, 2019.

\bibitem{antich2005extending}
J.~Antich and A.~Ortiz, ``Extending the potential fields approach to avoid
  trapping situations,'' in \emph{2005 IEEE/RSJ International Conference on
  Intelligent Robots and Systems}.\hskip 1em plus 0.5em minus 0.4em\relax IEEE,
  2005, pp. 1386--1391.

\bibitem{ge2005queues}
S.~S. Ge and C.-H. Fua, ``Queues and artificial potential trenches for
  multirobot formations,'' \emph{IEEE Transactions on Robotics}, vol.~21,
  no.~4, pp. 646--656, 2005.

\bibitem{vadakkepat2000evolutionary}
P.~Vadakkepat, K.~C. Tan, and W.~Ming-Liang, ``Evolutionary artificial
  potential fields and their application in real time robot path planning,'' in
  \emph{Proceedings of the 2000 congress on evolutionary computation. CEC00
  (Cat. No. 00TH8512)}, vol.~1.\hskip 1em plus 0.5em minus 0.4em\relax IEEE,
  2000, pp. 256--263.

\bibitem{orozco2019mobile}
U.~Orozco-Rosas, O.~Montiel, and R.~Sep{\'u}lveda, ``Mobile robot path planning
  using membrane evolutionary artificial potential field,'' \emph{Applied Soft
  Computing}, vol.~77, pp. 236--251, 2019.

\bibitem{koditschek1990robot}
D.~E. Koditschek and E.~Rimon, ``Robot navigation functions on manifolds with
  boundary,'' \emph{Advances in applied mathematics}, vol.~11, no.~4, pp.
  412--442, 1990.

\bibitem{masoud2007decentralized}
A.~A. Masoud, ``Decentralized self-organizing potential field-based control for
  individually motivated mobile agents in a cluttered environment: A
  vector-harmonic potential field approach,'' \emph{IEEE Transactions on
  Systems, Man, and Cybernetics-Part A: Systems and Humans}, vol.~37, no.~3,
  pp. 372--390, 2007.

\bibitem{panagou2015distributed}
D.~Panagou, D.~M. Stipanovi{\'c}, and P.~G. Voulgaris, ``Distributed
  coordination control for multi-robot networks using lyapunov-like barrier
  functions,'' \emph{IEEE Transactions on Automatic Control}, vol.~61, no.~3,
  pp. 617--632, 2015.

\bibitem{zhao2020multi}
T.~Zhao, H.~Li, and S.~Dian, ``Multi-robot path planning based on improved
  artificial potential field and fuzzy inference system,'' \emph{Journal of
  Intelligent \& Fuzzy Systems}, vol.~39, no.~5, pp. 7621--7637, 2020.

\bibitem{frew2005cooperative}
E.~Frew and D.~Lawrence, ``Cooperative stand-off tracking of moving targets by
  a team of autonomous aircraft,'' in \emph{AIAA Guidance, Navigation, and
  Control Conference and Exhibit}, 2005, p. 6363.

\bibitem{goncalves2010vector}
V.~M. Goncalves, L.~C. Pimenta, C.~A. Maia, B.~C. Dutra, and G.~A. Pereira,
  ``Vector fields for robot navigation along time-varying curves in $ n
  $-dimensions,'' \emph{IEEE Transactions on Robotics}, vol.~26, no.~4, pp.
  647--659, 2010.

\bibitem{frew2017tracking}
E.~W. Frew and D.~Lawrence, ``Tracking dynamic star curves using guidance
  vector fields,'' \emph{Journal of Guidance, Control, and Dynamics}, vol.~40,
  no.~6, pp. 1488--1495, 2017.

\bibitem{rezende2021constructive}
A.~M. Rezende, V.~M. Goncalves, and L.~C. Pimenta, ``Constructive time-varying
  vector fields for robot navigation,'' \emph{IEEE Transactions on Robotics},
  vol.~38, no.~2, pp. 852--867, 2021.

\bibitem{rezende2018robust}
A.~M. Rezende, V.~M. Gon{\c{c}}alves, G.~V. Raffo, and L.~C. Pimenta, ``Robust
  fixed-wing uav guidance with circulating artificial vector fields,'' in
  \emph{2018 IEEE/RSJ International Conference on Intelligent Robots and
  Systems (IROS)}.\hskip 1em plus 0.5em minus 0.4em\relax IEEE, 2018, pp.
  5892--5899.

\bibitem{pothen2017curvature}
A.~A. Pothen and A.~Ratnoo, ``Curvature-constrained lyapunov vector field for
  standoff target tracking,'' \emph{Journal of Guidance, Control, and
  Dynamics}, vol.~40, no.~10, pp. 2729--2736, 2017.

\bibitem{yao2021distributed}
W.~Yao, H.~G. de~Marina, Z.~Sun, and M.~Cao, ``Distributed coordinated path
  following using guiding vector fields,'' in \emph{2021 IEEE International
  Conference on Robotics and Automation (ICRA)}.\hskip 1em plus 0.5em minus
  0.4em\relax IEEE, 2021, pp. 10\,030--10\,037.

\bibitem{marchidan2020collision}
A.~Marchidan and E.~Bakolas, ``Collision avoidance for an unmanned aerial
  vehicle in the presence of static and moving obstacles,'' \emph{Journal of
  Guidance, Control, and Dynamics}, vol.~43, no.~1, pp. 96--110, 2020.

\bibitem{panagou2016distributed}
D.~Panagou, ``A distributed feedback motion planning protocol for multiple
  unicycle agents of different classes,'' \emph{IEEE Transactions on Automatic
  Control}, vol.~62, no.~3, pp. 1178--1193, 2016.

\bibitem{he2022simultaneous}
X.~He and Z.~Li, ``Simultaneous position and orientation planning of
  nonholonomic multi-robot systems: A dynamic vector field approach,''
  \emph{arXiv preprint arXiv:2209.00955}, 2022.

\bibitem{he2023novel}
X.~He, Z.~Sun, and Z.~Li, ``A novel vector-field-based motion planning for 3d
  nonholonomic robots,'' \emph{arXiv preprint arXiv:2302.11110}, 2023.

\bibitem{singh1996real}
L.~Singh, H.~Stephanou, and J.~Wen, ``Real-time robot motion control with
  circulatory fields,'' in \emph{Proceedings of IEEE International Conference
  on Robotics and Automation}, vol.~3.\hskip 1em plus 0.5em minus 0.4em\relax
  Minnesota: IEEE, 1996, pp. 2737--2742.

\bibitem{haddadin2011dynamic}
S.~Haddadin, R.~Belder, and A.~Albu-Sch{\"a}ffer, ``Dynamic motion planning for
  robots in partially unknown environments,'' \emph{IFAC Proceedings Volumes},
  vol.~44, no.~1, pp. 6842--6850, 2011.

\bibitem{ataka2018reactive}
A.~Ataka, H.-K. Lam, and K.~Althoefer, ``Reactive magnetic-field-inspired
  navigation for non-holonomic mobile robots in unknown environments,'' in
  \emph{2018 IEEE International Conference on Robotics and Automation
  (ICRA)}.\hskip 1em plus 0.5em minus 0.4em\relax Australia: IEEE, 2018, pp.
  6983--6988.

\bibitem{laha2021reactive}
R.~Laha, L.~F. Figueredo, J.~Vrabel, A.~Swikir, and S.~Haddadin, ``Reactive
  cooperative manipulation based on set primitives and circular fields,'' in
  \emph{2021 IEEE International Conference on Robotics and Automation
  (ICRA)}.\hskip 1em plus 0.5em minus 0.4em\relax Xi'an: IEEE, 2021, pp.
  6577--6584.

\bibitem{becker2021circular}
M.~Becker, T.~Lilge, M.~A. M{\"u}ller, and S.~Haddadin, ``Circular fields and
  predictive multi-agents for online global trajectory planning,'' \emph{IEEE
  Robotics and Automation Letters}, vol.~6, no.~2, pp. 2618--2625, 2021.

\bibitem{ames2019control}
A.~D. Ames, S.~Coogan, M.~Egerstedt, G.~Notomista, K.~Sreenath, and P.~Tabuada,
  ``Control barrier functions: Theory and applications,'' in \emph{2019 18th
  European control conference (ECC)}.\hskip 1em plus 0.5em minus 0.4em\relax
  IEEE, 2019, pp. 3420--3431.

\bibitem{lamport1977proving}
L.~Lamport, ``Proving the correctness of multiprocess programs,'' \emph{IEEE
  Transactions on Software Engineering}, no.~2, pp. 125--143, 1977.

\bibitem{slotine1991applied}
J.-J.~E. Slotine, W.~Li \emph{et~al.}, \emph{Applied nonlinear control}.\hskip
  1em plus 0.5em minus 0.4em\relax Prentice hall Englewood Cliffs, NJ, 1991,
  vol. 199, no.~1.

\bibitem{abraham2012manifolds}
R.~Abraham, J.~E. Marsden, and T.~Ratiu, \emph{Manifolds, tensor analysis, and
  applications}.\hskip 1em plus 0.5em minus 0.4em\relax Springer Science \&
  Business Media, 2012, vol.~75.

\bibitem{ames2014rapidly}
A.~D. Ames, K.~Galloway, K.~Sreenath, and J.~W. Grizzle, ``Rapidly
  exponentially stabilizing control lyapunov functions and hybrid zero
  dynamics,'' \emph{IEEE Transactions on Automatic Control}, vol.~59, no.~4,
  pp. 876--891, 2014.

\bibitem{freeman2008robust}
R.~Freeman and P.~V. Kokotovic, \emph{Robust nonlinear control design:
  state-space and Lyapunov techniques}.\hskip 1em plus 0.5em minus 0.4em\relax
  Springer Science \& Business Media, 2008.

\bibitem{borrmann2015control}
U.~Borrmann, L.~Wang, A.~D. Ames, and M.~Egerstedt, ``Control barrier
  certificates for safe swarm behavior,'' \emph{IFAC-PapersOnLine}, vol.~48,
  no.~27, pp. 68--73, 2015.

\bibitem{ames2014control}
A.~D. Ames, J.~W. Grizzle, and P.~Tabuada, ``Control barrier function based
  quadratic programs with application to adaptive cruise control,'' in
  \emph{53rd IEEE Conference on Decision and Control}.\hskip 1em plus 0.5em
  minus 0.4em\relax IEEE, 2014, pp. 6271--6278.

\bibitem{wang2016multi}
L.~Wang, A.~D. Ames, and M.~Egerstedt, ``Multi-objective compositions for
  collision-free connectivity maintenance in teams of mobile robots,'' in
  \emph{2016 IEEE 55th Conference on Decision and Control (CDC)}.\hskip 1em
  plus 0.5em minus 0.4em\relax IEEE, 2016, pp. 2659--2664.

\bibitem{glotfelter2017nonsmooth}
P.~Glotfelter, J.~Cort{\'e}s, and M.~Egerstedt, ``Nonsmooth barrier functions
  with applications to multi-robot systems,'' \emph{IEEE Control Systems
  Letters}, vol.~1, no.~2, pp. 310--315, 2017.

\bibitem{wang2017safe}
L.~Wang, A.~D. Ames, and M.~Egerstedt, ``Safe certificate-based maneuvers for
  teams of quadrotors using differential flatness,'' in \emph{2017 IEEE
  International Conference on Robotics and Automation (ICRA)}.\hskip 1em plus
  0.5em minus 0.4em\relax IEEE, 2017, pp. 3293--3298.

\bibitem{squires2018constructive}
E.~Squires, P.~Pierpaoli, and M.~Egerstedt, ``Constructive barrier certificates
  with applications to fixed-wing aircraft collision avoidance,'' in \emph{2018
  IEEE Conference on Control Technology and Applications (CCTA)}.\hskip 1em
  plus 0.5em minus 0.4em\relax IEEE, 2018, pp. 1656--1661.

\bibitem{squires2021safety}
E.~Squires, R.~Konda, P.~Pierpaoli, S.~Coogan, and M.~Egerstedt, ``Safety with
  limited range sensing constraints for fixed wing aircraft,'' in \emph{2021
  IEEE International Conference on Robotics and Automation (ICRA)}.\hskip 1em
  plus 0.5em minus 0.4em\relax IEEE, 2021, pp. 9065--9071.

\bibitem{AIRBUS}
Airbus, ``Airbus skyways: the future of the parcel delivery in smart cities,''
  2019, \url{https://www.embention.com/project/airbus-parcel-delivery/}.

\bibitem{mao2022making}
P.~Mao and Q.~Quan, ``Making robotics swarm flow more smoothly: A regular
  virtual tube model,'' in \emph{2022 IEEE/RSJ International Conference on
  Intelligent Robots and Systems (IROS)}.\hskip 1em plus 0.5em minus
  0.4em\relax IEEE, 2022, pp. 4498--4504.

\bibitem{mao2023optimal}
P.~Mao, R.~Fu, and Q.~Quan, ``Optimal virtual tube planning and control for
  swarm robotics,'' \emph{arXiv preprint arXiv:2304.11407}, 2023.

\bibitem{song2023speed}
W.~Song, Y.~Gao, and Q.~Quan, ``Speed and density planning for a
  speed-constrained robot swarm through a virtual tube,'' \emph{arXiv preprint
  arXiv:2310.00623}, 2023.

\bibitem{quan2022far}
Q.~Quan, R.~Fu, and K.-Y. Cai, ``How far two uavs should be subject to
  communication uncertainties,'' \emph{IEEE Transactions on Intelligent
  Transportation Systems}, vol.~24, no.~1, pp. 429--445, 2022.

\bibitem{gao2023non}
Y.~Gao, C.~Bai, R.~Fu, and Q.~Quan, ``A non-potential orthogonal vector field
  method for more efficient robot navigation and control,'' \emph{Robotics and
  Autonomous Systems}, vol. 159, p. 104291, 2023.

\bibitem{lv2023autonomous}
S.~Lv, Y.~Gao, J.~Che, and Q.~Quan, ``Autonomous drone racing: Time-optimal
  spatial iterative learning control within a virtual tube,'' in \emph{2023
  IEEE International Conference on Robotics and Automation (ICRA)}.\hskip 1em
  plus 0.5em minus 0.4em\relax IEEE, 2023, pp. 3197--3203.

\end{thebibliography}

\begin{IEEEbiography}[{\includegraphics[width=1in,height=1.25in,clip,keepaspectratio]{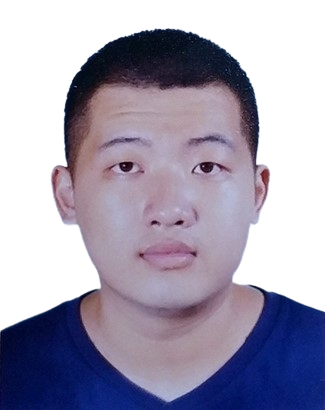}}]{Yan Gao} received the B.S. degree in control science and engineering from Harbin Institute of Technology, Harbin, China, in 2017. He is currently pursuing Ph.D. in control science and engineering with the School of Automation Science and Electrical Engineering, Beihang University, Beijing, China. His main research interests include UAV swarm and quadcopter control.
\end{IEEEbiography}

\begin{IEEEbiography}[{\includegraphics[width=1in,height=1.25in,clip,keepaspectratio]{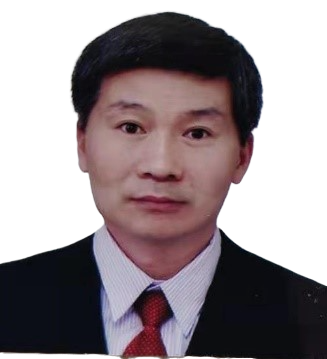}}]{Chenggang Bai} received the MSc degree in statistics from the Nanjing University of Aeronautics and Astronautics, China, in 1990, and the PhD degree in control theory and control engineering from Zhejiang University, China. In November 2001, he joined the faculty of the School of Automation Science and Electrical Engineering, Beihang University, China, where he has been a professor since July 2009. His research interests include reliable flight control, software reliability, and software testing.
\end{IEEEbiography}

\begin{IEEEbiography}[{\includegraphics[width=1in,height=1.25in,clip,keepaspectratio]{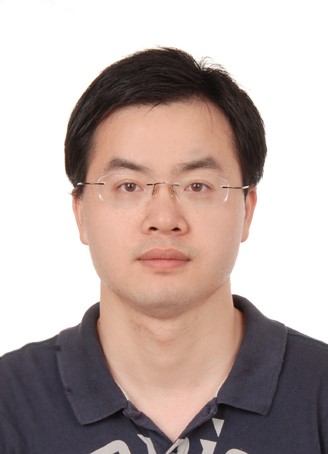}}]{Quan Quan}received the B.S. and Ph.D. degrees in control science and engineering from Beihang University, Beijing, China, in 2004, and 2010, respectively. Since 2022, he has been a Professor with Beihang University in control science and engineering, where he is currently with the School of Automation Science and Electrical Engineering. His research interests include reliable flight control, swarm intelligence, vision-based navigation, and health assessment.
\end{IEEEbiography}

\end{document}